\pdfoutput=1

\documentclass[11pt]{article}

\usepackage{ACL2023}

\usepackage{times}
\usepackage{latexsym}

\usepackage[T1]{fontenc}

\usepackage[utf8]{inputenc}

\usepackage{microtype}

\usepackage{inconsolata}

\usepackage{algorithm}
\usepackage{algpseudocode}
\renewcommand{\algorithmicrequire}{\textbf{Input:}}
\renewcommand{\algorithmicensure}{\textbf{Output:}}
\makeatletter
\algnewcommand{\LineComment}[1]{\Statex \hskip\ALG@thistlm \(\triangleright\) #1}
\makeatother

\usepackage{float} %
\usepackage{booktabs}       %
\usepackage{makecell}
\usepackage{array}
\newcolumntype{L}[1]{>{\raggedright\let\newline\\\arraybackslash\hspace{0pt}}m{#1}}
\newcolumntype{C}[1]{>{\centering\let\newline\\\arraybackslash\hspace{0pt}}m{#1}}
\newcolumntype{R}[1]{>{\raggedleft\let\newline\\\arraybackslash\hspace{0pt}}m{#1}}
\usepackage{amsfonts}       %
\usepackage{nicefrac}       %
\usepackage{graphicx}
\usepackage{xspace}
\usepackage{amsmath}
\usepackage[inline]{enumitem}
\usepackage[noabbrev]{cleveref}
\usepackage{changepage}
\usepackage{ulem,comment,multirow}
\usepackage{makecell}
\usepackage{amssymb}
\usepackage{marvosym}
\usepackage{wrapfig}
\usepackage[section]{placeins}
\usepackage{subcaption}
\newcommand{\footurl}[1]{\footnote{\url{#1}}}

\newcommand{\langcode}[1]{\texttt{#1}}

\DeclareMathOperator*{\argmin}{argmin}

\makeatletter
\DeclareRobustCommand\onedot{\futurelet\@let@token\@onedot}
\def\@onedot{\ifx\@let@token.\else.\null\fi\xspace}

\def\wrt{w.r.t\onedot} 

\makeatother

\newcolumntype{H}{>{\setbox0=\hbox\bgroup}c<{\egroup}@{}}
\newcolumntype{P}[1]{>{\centering\arraybackslash}p{#1}}

\newcommand{\ablation}{MMT dataset}
\newcommand{\flores}{{\textsc{Flores-200}}}

\newcommand{\modHiddenDim}{1024}
\newcommand{\modFFNDim}{8192}
\newcommand{\modAttHeads}{16}
\newcommand{\modLayers}{24}
\newcommand{\moeExperts}{64}

\DeclareMathOperator\moe{MoE}

\newcommand{\topgating}[1]{Top-#1-Gating}
\DeclareMathOperator\softmax{softmax}
\newcommand{\lang}{{{<}\ell{>}}}

\newcommand{\y}{\boldsymbol y}

\newcommand{\ffn}[1]{\operatorname{FFN}_{#1}}

\newcommand{\clupdates}{U}
\newcommand{\sgmoelong}{Sparsely-Gated Mixture-of-Experts (MoE)}

\newcommand{\moetokendropout}{MoE Expert Output Masking}
\newcommand{\moetokendropoutabbrv}{\textsc{EOM}}
\newcommand{\fom}{\textsc{FOM}}
\newcommand{\clsr}{Conditional MoE Routing}
\newcommand{\clsrabbrv}{\textsc{CMR}}
\newcommand{\clsrdrop}{p_\text{cmr}}
\newcommand{\clsrbudget}{b}
\newcommand{\moetokdrop}{p_\text{eom}}
\newcommand{\fomdrop}{p_\text{fom}}
\newcommand{\drop}{p_\text{drop}}
\newcommand{\localgatedropout}{Gating Dropout}
\newcommand{\chrf}{chrF\nolinebreak\hspace{-.05em}\raisebox{.4ex}{\tiny\bf +}\nolinebreak\hspace{-.10em}\raisebox{.4ex}{\tiny\bf +}}

\newcommand{\modelsize}{1.3B}
\newcommand{\modelsizesm}{615M}

\usepackage{tikz, pgfplots}
\usepackage{pgffor,pgfmath}
\usepackage{xparse}
\usepackage{ifthen}
\usepackage{pgfplotstable}
\usepackage{pdftexcmds}
\makeatletter
\newcommand{\strequal}[2]{\pdf@strcmp{#1}{#2}==0}
\makeatother

\definecolor{lightgray}{rgb}{.95,.95,.95}      %
\definecolor{dimgray}{rgb}{.35,.35,.35}      %
\definecolor{darkgray}{rgb}{.20,.20,.20}     %

\tikzset{
    txt/.style={
      rounded corners=1pt,
      text=darkgray,
      font=\small,
      align=center,
      inner sep=1pt,
      outer sep=2pt,
      minimum height=5mm,
    },
    vector/.style={ %
      draw=dimgray,
      rounded corners=1pt,
      text=darkgray,
      inner sep=2pt,
      minimum height=20pt,
      minimum width=4pt, 
      line width=1pt,
      font=\small,
      align=center
    },
    neurone/.style={ %
      draw=dimgray,
      text=darkgray,
      inner sep=2pt,
      minimum height=6pt, 
      minimum width=6pt,
      line width=.4pt,
      font=\small,
      align=center
    },
    conn/.style={
      -{Straight Barb[angle=60:1pt 2]}, line width=1.0pt,
        darkgray
    },
    block/.style={
      rectangle,draw=dimgray,
      rounded corners=1pt,
      text=darkgray,
      inner sep=2pt,
      minimum height=2mm,
      minimum width=2mm,
      node distance=4pt and 10pt,
      line width=0.5pt,
      font=\small,
      align=center
    },
    layer/.style={
      rectangle,draw=dimgray,
      rounded corners=2pt,
      text=darkgray,
      inner sep=2pt,
      minimum height=4mm,
      minimum width=3.5cm,
      node distance=4pt and 10pt,
      line width=0.8pt,
      font=\small,
      align=center
    },
    layernorm/.style={
      fill=Set1B!20,
      rectangle,draw=dimgray,
      rounded corners=2pt,
      text=darkgray,
      inner sep=2pt,
      minimum height=4mm,
      minimum width=3.5cm,
      node distance=4pt and 10pt,
      line width=0.5pt,
      font=\small,
      align=center
    },
    attnlayer/.style={
      fill=Set1A!20,
      rectangle,draw=dimgray,
      rounded corners=2pt,
      text=darkgray,
      inner sep=2pt,
      minimum height=4mm,
      minimum width=3.5cm,
      node distance=4pt and 10pt,
      line width=0.5pt,
      font=\small,
      align=center
    },
    ffnlayer/.style={
      fill=Set1C!20,
      rectangle,draw=dimgray,
      rounded corners=2pt,
      text=darkgray,
      inner sep=2pt,
      minimum height=4mm,
      minimum width=3.5cm,
      node distance=4pt and 10pt,
      line width=0.5pt,
      font=\small,
      align=center
    },
    residual/.style={
        circle, draw=darkgray,minimum width=8pt,line width=1.0pt,
        path picture={
            \draw[darkgray]
            (path picture bounding box.south) -- (path picture bounding box.north) (path picture bounding box.west) -- (path picture bounding box.east);
    }}
}

\pgfplotsset{
    compat=1.17,
    grid style={darkgray},
    minor grid style={dimgray!20},
    major grid style={dimgray!20},
    axis line style = { darkgray }, 
    every axis plot/.append style={line width=1.5pt, mark options=solid, mark size=4pt},
    legend style={draw = darkgray, rounded corners=0pt, fill = white, font=\Large},
    legend cell align={left},
    legend image code/.code={%
        \draw[mark repeat=2,mark phase=2]
        plot coordinates {
            (0cm,0cm)
            (0.2cm,0cm)
            (0.4cm,0cm)
        };%
    },
    custom ybar legend/.style={%
        legend image code/.code={
            \fill [##1] (0pt, -3pt) rectangle +(3pt,8pt);
        },
    },
    tick style ={color = dimgray!30 },
    tick label style={font=\normalsize},
    label style={font=\normalsize},
}

\usetikzlibrary{decorations, shapes, arrows, arrows.meta, positioning, decorations.pathreplacing, calc}
\usetikzlibrary{pgfplots.groupplots, patterns}
\usetikzlibrary{pgfplots.statistics}
\usepgfplotslibrary{external}
\usepgfplotslibrary{colorbrewer}
\pgfplotsset{cycle list/Dark2}

\pgfplotsset{%
    discard if out of range/.style n args={3}{%
        x filter/.code={%
            \edef\tempa{\thisrow{#1}}
            \edef\tempb{#2}
            \edef\tempc{#3}
            \ifdim\tempa pt> \tempb pt
              \ifdim\tempa pt< \tempc pt
              \else
                
              \fi
            \else
              
            \fi
        }
    }
}  

\pgfplotsset{
    discard if not/.style 2 args={
        x filter/.code={
            \edef\tempa{\thisrow{#1}}
            \edef\tempb{#2}
            \ifx\tempa\tempb
            \else
                
            \fi
        }
    }
}

\definecolor{Dark2A}{RGB}{27,158,119}
\definecolor{Dark2B}{RGB}{217,95,2}
\definecolor{Dark2C}{RGB}{117,112,179}
\definecolor{Dark2D}{RGB}{231,41,138}
\definecolor{Dark2H}{RGB}{102,166,30}
\definecolor{Dark2F}{RGB}{230,171,2}
\definecolor{Dark2G}{RGB}{166,118,29}
\definecolor{Dark2E}{RGB}{102,102,102}

\definecolor{Set1AP}{HTML}{a10303} %
\definecolor{Set1BP}{HTML}{0614d6} %
\definecolor{Set1CP}{HTML}{299e3f} %

\definecolor{Set1A}{HTML}{CC3311} %
\definecolor{Set1B}{HTML}{377EB2} %
\definecolor{Set1C}{HTML}{228833} %
\definecolor{Set1D}{HTML}{DDAA33} %
\definecolor{Set1E}{HTML}{AA4499} %
\definecolor{Set1F}{HTML}{EE7733} %
\definecolor{Set1G}{HTML}{6699CC} %
\definecolor{Set1H}{HTML}{A6A6A6} %
\definecolor{Set1I}{HTML}{117733} %
\definecolor{Set1J}{HTML}{882255} %
\definecolor{Set1K}{HTML}{009988} %
\definecolor{Set1L}{HTML}{EE3377} %
\definecolor{Set1M}{HTML}{997700} %
\definecolor{Set1N}{HTML}{EE6677} %
\definecolor{Set1O}{HTML}{EEE966} %
\definecolor{Set1P}{HTML}{6647FF} %
\definecolor{Set1Q}{HTML}{0FF707} %
\definecolor{Set1R}{HTML}{07F7Eb} %
\definecolor{Set1S}{HTML}{014A46} %
\definecolor{Set1T}{HTML}{822203} %
\definecolor{Set1U}{HTML}{5E8203} %
\definecolor{Set1V}{HTML}{b50951} %
\definecolor{Set1X}{HTML}{ca58ed} %
\definecolor{Set1Y}{HTML}{453c80} %
\definecolor{Set1Z}{HTML}{82947b} %

\definecolor{IndoEuropean}{RGB}{154, 227, 245}
\definecolor{Uralic}{RGB}{89, 180, 240}
\definecolor{AfroAsiatic}{RGB}{56, 119, 161}
\definecolor{AtlanticCongo}{RGB}{182, 215, 168} 
\definecolor{Mande}{RGB}{117, 191, 86}
\definecolor{Nilotic}{RGB}{65, 145, 32}
\definecolor{Saharan}{RGB}{29, 97, 1}
\definecolor{Austronesian}{RGB}{252, 229, 205}
\definecolor{AustroAsiatic}{RGB}{255, 200, 143}
\definecolor{TaiKadai}{RGB}{255, 142, 25}
\definecolor{SinoTibetan}{RGB}{250, 224, 30}
\definecolor{Japonic}{RGB}{208, 175, 250}
\definecolor{Koreanic}{RGB}{155, 86, 240}
\definecolor{Turkic}{RGB}{252, 187, 201}
\definecolor{Dravidian}{RGB}{250, 97, 131}
\definecolor{Mongolic}{RGB}{247, 17, 68}
\definecolor{Kartvelian}{RGB}{247, 119, 147}
\definecolor{Basque}{RGB}{91, 15, 0} 
\definecolor{Constructed}{RGB}{13, 13, 13}
\definecolor{Aymaran}{RGB}{204, 204, 204}  
\definecolor{Quechuan}{RGB}{153, 153, 153}
\definecolor{Tupian}{RGB}{67, 67, 67}

\definecolor{Latin}{RGB}{154, 227, 245}
\definecolor{Greek}{RGB}{89, 180, 240}
\definecolor{Georgian}{RGB}{56, 119, 161}
\definecolor{Armenian}{RGB}{19, 95, 145}
\definecolor{Cyrillic}{RGB}{8, 43, 133}

\definecolor{Bengali}{RGB}{169, 250, 135}
\definecolor{Devanagari}{RGB}{137, 252, 88}
\definecolor{Gujarati}{RGB}{84, 237, 19}
\definecolor{Gurmukhi}{RGB}{49, 156, 3}
\definecolor{Oriya}{RGB}{39, 94, 15}
\definecolor{Tibetan}{RGB}{65, 99, 51}

\definecolor{Sinhala}{RGB}{242, 241, 199}
\definecolor{Malayalam}{RGB}{247, 245, 126}
\definecolor{Tamil}{RGB}{252, 248, 48}
\definecolor{Telugu}{RGB}{196, 192, 2}
\definecolor{Kannada}{RGB}{156, 152, 12}

\definecolor{Myanmar}{RGB}{247, 202, 156}
\definecolor{Lao}{RGB}{247, 170, 92}
\definecolor{Khmer}{RGB}{247, 136, 22}
\definecolor{Thai}{RGB}{156, 80, 3}

\definecolor{Japanese}{RGB}{208, 175, 250}
\definecolor{Hangul}{RGB}{111, 20, 227}
\definecolor{HanT}{RGB}{252, 187, 201}
\definecolor{HanS}{RGB}{250, 97, 131}
\definecolor{Arabic}{RGB}{204, 204, 204}  
\definecolor{Hebrew}{RGB}{153, 153, 153}
\definecolor{Tifinagh}{RGB}{89, 13, 4}
\definecolor{Geez}{RGB}{255, 0, 0}

\pgfplotscreateplotcyclelist{CustomListWithMarkers}{%
    smooth, Set1A, mark=x, mark size=2.5pt, line width=1.pt\\
    smooth, Set1B, mark=o, mark size=2.5pt, line width=1.pt\\
    smooth, Set1C, mark=triangle, mark size=2.5pt, line width=1.pt\\
    smooth, Set1D, mark=+, mark size=2.5pt, line width=1.pt\\
    smooth, Set1E, mark=square, mark size=1.9pt, line width=1.pt\\
    smooth, Set1F, mark=pentagon, mark size=2.5pt, line width=1.pt\\
}

\pgfplotscreateplotcyclelist{CustomList}{%
    smooth, Set1A, line width=1pt\\  
    smooth, Set1B, line width=1pt\\
    smooth, Set1C, line width=1pt\\
    smooth, Set1D, line width=1pt\\
    smooth, Set1E, line width=1pt\\  
    smooth, Set1F, line width=1pt\\
}

\pgfplotscreateplotcyclelist{CustomListBar}{%
    smooth, draw opacity=0, fill=Set1H, line width=0.5pt\\
    smooth, draw opacity=0, fill=Set1C, line width=0.5pt\\
    smooth, draw opacity=0, fill=Set1B, line width=0.5pt\\
    smooth, draw opacity=0, fill=Set1A, line width=0.5pt\\
    smooth, draw opacity=0, fill=Set1D, line width=0.5pt\\
    smooth, draw opacity=0, fill=Set1E, line width=0.5pt\\
}

\pgfplotscreateplotcyclelist{CustomListWithMarkersEvaluation}{%
    smooth, Set1A, mark=o, mark size=1.5pt, line width=1pt\\% nllb-200 xx-eng
    smooth, Set1B, mark=o, mark size=1.5pt, line width=1pt\\% nllb-200 baseline xx-eng
    smooth, Set1H, mark=o, mark size=1.5pt, line width=1pt\\% m2m-100 xx-eng
    smooth, Set1A, mark=x, mark size=2.5pt, line width=1pt\\% nllb-200
    smooth, Set1B, mark=x, mark size=2.5pt, line width=1pt\\% nllb-200 baseline
    smooth, Set1H, mark=x, mark size=2.5pt, line width=1pt\\% m2m-100
    smooth, Set1D, mark=x, mark size=2.5pt, line width=1pt\\% nllb-125
    smooth, Set1C, mark=x, mark size=2.5pt, line width=1pt\\% wmt
}

\NewDocumentCommand{\plotlangPPL}{ O{\empty} O{\empty} m }{%
\begin{tikzpicture}
    \begin{axis}[
        grid=both,
        xmin=10000,xmax=100000,
        minor tick num=1,
        xlabel=#1,
        xtick={10000,40000,...,100000},
        xticklabels={10k,40k,70k,100k},
        y tick label style={font=\tiny, xshift=2pt},
        x tick label style={font=\tiny, yshift=1pt},
        label style={font=\scriptsize},
        scaled x ticks=false,
        legend style={at={(0.99,0.99)},anchor=north east, font=\tiny},
        cycle list name=CustomListWithMarkers,
        height=4.5cm,
        width=5.5cm,
        ylabel=#2,
       ]
\foreach \model/\label in {%
    dense_drop0/Dense-drop0.0,%
    dense_drop0.3/Dense-drop0.3,%
    moe64_drop0/MoE64-drop0.0,%
    moe64_drop0.3/MoE64-drop0.3,%
    moe64_drop0.3_tok0.2/MoE64-drop0.3-tok0.2%
    }{
    \addplot+
    table [y=#3,x=updates]{results/ablation/\model.dat};
}
\end{axis}
\end{tikzpicture}}

\newif\ifcomments
\ifcomments
    \providecommand{\annote}[3]{{\color{#3}%
            \colorbox{#3}{\bfseries\sffamily\tiny\textcolor{white}{#2}}
            \color{#3}
            $\blacktriangleright$\footnotesize\emph{#1}$\blacktriangleleft$}%
    }
    \providecommand{\todo}[1]{\annote{#1}{TODO}{red}}
    \providecommand{\sannote}[2]{{\color{#2}
            $\blacktriangleright$\footnotesize\emph{#1}$\blacktriangleleft$}%
    }
    \providecommand{\stodo}[1]{\sannote{#1}{red}}
    \providecommand\maha[1]{[\textcolor{Set1BP}{Maha: {#1}}]}
    \providecommand\anna[1]{[\textcolor{Set1E}{Anna: {#1}}]}
    \providecommand\shruti[1]{[\textcolor{Set1CP}{Shruti: {#1}}]}

\else
    \providecommand{\maha}[1]{}
    \providecommand{\anna}[1]{}
    \providecommand{\shruti}[1]{}

    \providecommand{\annote}[3]{}
    \providecommand{\todo}[1]{}
    \providecommand{\sannote}[2]{}
    \providecommand{\stodo}[1]{}

\fi

\title{Fixing MoE Over-Fitting on Low-Resource Languages in Multilingual Machine Translation}
\author{
  Maha Elbayad\Thanks{~Equal contribution}\\
  Meta AI\\
  \texttt{elbayadm@meta.com}
  \And
  Anna Sun$^*$\\
  Meta AI\\
  \texttt{annaysun@meta.com}
  \And
  Shruti Bhosale\\
  Meta AI\\
  \texttt{shru@meta.com}
}

\begin{document}

\maketitle

\begin{abstract}%
    Sparsely gated Mixture of Experts (MoE) models have been shown to be a compute-efficient method to scale model capacity for multilingual machine translation. However, for low-resource tasks, MoE models severely over-fit. We show effective regularization strategies, namely dropout techniques for MoE layers in EOM and FOM, Conditional MoE Routing and Curriculum Learning methods that prevent over-fitting and improve the performance of MoE models on low-resource tasks without adversely affecting high-resource tasks. On a massively multilingual machine translation benchmark, our strategies result in about +1 \chrf{} improvement in very low resource language pairs. We perform an extensive analysis of the learned MoE routing to better understand the impact of our regularization methods and how we can improve them.
\end{abstract}

\section{Introduction}
\label{sec:intro}
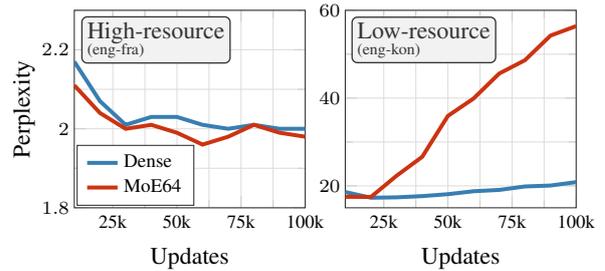
\begin{figure}[t]
    \begin{tikzpicture}[scale=1]
\begin{groupplot}[
        group style={group size=2 by 1, horizontal sep=15pt, vertical sep=20pt},
        grid=both,
        minor tick num=1,
        label style={font=\small},
        y label style={yshift=-5pt},
        y tick label style={font=\scriptsize, xshift=2pt},
        x tick label style={font=\scriptsize, yshift=1pt},
        scaled x ticks=false,
        y axis line style=-,
        height=4.2cm,
        width=0.6*\columnwidth,
        xtick={25000,50000,...,100000},
        xticklabels={25k,50k,75k,100k},
        xmin=10000,xmax=100000,
       ]
\nextgroupplot[
    ymin=1.8,ymax=2.3,
    xlabel={Updates},
    ylabel={Perplexity},
    legend style={at={(0.01,0.01)},anchor=south west, font=\small, nodes={scale=0.8, transform shape}}
   ]

   \addplot+[Set1B]
    table [y=eng-fra,x=updates]{results/ablation/dense_drop0.dat};
    \addlegendentryexpanded{Dense}

    \addplot+[Set1A]
    table [y=eng-fra,x=updates]{results/ablation/moe64_drop0.dat};
    \addlegendentryexpanded{MoE64}

\nextgroupplot[
    ymin=15,ymax=60,
    xlabel={Updates},
   ]
   \addplot+[Set1B]
    table [y=eng-kon,x=updates]{results/ablation/dense_drop0.dat};
    \addlegendentryexpanded{Dense}

        \addplot+[Set1A]
    table [y=eng-kon,x=updates]{results/ablation/moe64_drop0.dat};
    \addlegendentryexpanded{MoE64}
    \legend{}
\end{groupplot}
\node[layer, draw=darkgray, anchor=west, 
    align=left, font=\small, line width=0.5pt,
    fill=lightgray, text=darkgray, minimum width=1cm
    ]  at (rel axis cs: 1.2, 0.85) {Low-resource\\[-4pt]\tiny (eng-kon)};
\node[layer, draw=darkgray, anchor=west, 
    align=left, font=\small, line width=0.5pt,
    fill=lightgray, text=darkgray, minimum width=1cm
    ]  at (rel axis cs: 0.03, 0.85) {High-resource\\[-4pt]\tiny (eng-fra)};
\end{tikzpicture}
    \vspace{-18pt}
    \caption{Validation perplexity of dense and MoE (64 experts) models. We show a high-resource direction that does not suffer from over-fitting, when a low resource direction sees extreme over-fitting.
    }
    \label{fig:intro:overfit}
\end{figure}

Training massively multitask models such as multilingual machine translation models benefit from transfer learning across different tasks. But they also suffer from reduced model capacity per task and potential interference between conflicting tasks. 
Scaling up models has been shown to be a very effective strategy in many natural language processing tasks such as language modeling, massively multilingual translation and natural language understanding \citep{brown2020language,kaplan2020scaling}. Most of these advancements have focused on training increasingly larger dense models.
However, dense model scaling is computationally expensive, as a result, various sparse model architectures have been proposed to increase model capacity without incurring additional compute costs; the most commonly used one is the \sgmoelong{} layer ~\citep{shazeer2017outrageously,lepikhin2020gshard,du2021glam,hwang2022tutel,zoph2022designing}.

MoE models are a type of conditional compute models~\citep{bengio2013conditional,almahairi:2016:dcn} that activate a subset of model parameters per input, as opposed to \textit{dense} models that activate all model parameters. 
MoE models unlock significant representational capacity while maintaining the same inference and training efficiencies in terms of FLOPs as compared to the core dense architecture. 
As a result, past work has demonstrated improved performance on multitask models such as multilingual machine translation when using MoE models~\citep{lepikhin2020gshard, kim2021zcode, fedus2022switch, zoph2022designing}.

But we notice that, on imbalanced datasets, MoE models suffer from over-fitting on low resource tasks i.e.,  tasks with relatively less training data. \Cref{fig:intro:overfit} illustrates this phenomenon on a multilingual translation benchmark. We see that 
\langcode{eng-fra}, a high-resource translation direction, does not over-fit with either dense or MoE models. 
On the other hand,
\langcode{eng-kon}, a low-resource translation direction, extremely over-fits with the MoE model compared to the dense model. 

In this work, we introduce four effective strategies to reduce the over-fitting of MoE models on low-resource tasks in a massively multilingual MT benchmark: 
\begin{enumerate}
\itemsep0pt
\item Dropout techniques for MoE layers: we introduce Expert Output Masking (\moetokendropoutabbrv{}) and Final Output Masking (FOM), two dropout methods specific to MoE layers that we apply on top of overall  dropout.
\item Conditional MoE Routing (\clsrabbrv{}): We train an additional gate to decide when to route a token to an MoE layer vs. a shared dense layer.
\item Curriculum Learning (CL): We introduce low-resource pairs that are prone to over-fitting in the later stages of model training.
\end{enumerate}
On a massively multilingual MT benchmark,\footnote{53 languages and 110 translation directions and approximately 1.7B training examples}
we experimentally demonstrate the effectiveness of each of these strategies.
Particularly, we observe close to +1 \chrf{} improvements with EOM, FOM, CMR and CL strategies on very low resource language directions out of English.

\section{Background}\label{sec:modeling:prelim}
We first describe the multilingual machine translation (MMT) task setup, the dense backbone architecture, and how we augment it with MoE layers. 

\paragraph{Multilingual Machine Translation.} 
We model multilingual neural machine translation as a sequence-to-sequence task, where we condition on an input sequence in the source language with an encoder and generate the output sequence in the expected target language with a decoder~\citep{sutskever14nips}. We train to maximize the probability of the translation sequence in the target language given the source sequence, in addition to the source language $\ell_s$ and the target language $\ell_t$.

\paragraph{Model Architecture.}

Our sequence-to-sequence multilingual machine translation model is based on the Transformer encoder-decoder architecture~\citep{vaswani2017attention}. 

To prime the model for multilingual translation, we prefix the source sequence with the source language $\ell_s$ and the target sequence with the target language $\ell_t$.

\paragraph{Sparsely Gated Mixture of Experts.} 
In both Transformer encoder and decoder, we replace every other dense FFN sublayer with an MoE sublayer.
The MoE sublayer consists of $E$ feed-forward networks (FFN), denoted with $(\ffn{1},\ffn{2},\ldots,\ffn{E})$. 
A gating network, consisting of a softmax-normalized linear layer with weights $W_g$, is attached to each MoE sublayer to decide how to route tokens to experts. Given an input token $x_t$ the output of the MoE sublayer is evaluated as:
\begin{align}
&\mathcal G_{t} = \text{\topgating{k}}(\softmax(W_g \cdot x_t)),\label{eqn:topkgating}\\ %
&\moe(x_t) = \sum\limits_{e=1}^E \mathcal G_{te}\cdot \ffn{e}(x_t),
\label{eqn:moe-layer}
\end{align}
with $\mathcal G_{t}\in\mathbb R^E$ the routing vector computed by the gating network, i.e.,  for each expert, $\mathcal G_{t,e}$ is the contribution of the e$^{\text{th}}$ expert ($\ffn e$) in the MoE output.
We follow the \topgating{k} algorithm of~\citet{lepikhin2020gshard} and dispatch each token to at most $k{=}2$ experts. 

The sparse MoE model learns to route input tokens to the corresponding top-2 experts by optimizing a linearly weighted combination of label-smoothed cross entropy, $L_{\text{MT}}$,~($\epsilon{=}0.1$, \citet{szegedy:inception:2015}) and an auxiliary load balancing loss, $L_{\text{MoE}}$~\citep{shazeer2017outrageously},
\begin{align}
L = L_{\text{MT}} + \lambda_{\text{MoE}} L_{\text{MoE}}.
\label{eqn:moeloss}
\end{align}
This additional loss term ($L_{\text{MoE}}$) pushes the tokens to be uniformly distributed across experts.
We set $\lambda_{\text{MoE}}$ to 0.01 in all our experiments. We refer the reader to~\citet{lepikhin2020gshard} for more on the optimization of MoE models.

\section{Fixing over-fitting on low-resource tasks}
\label{sec:methods}
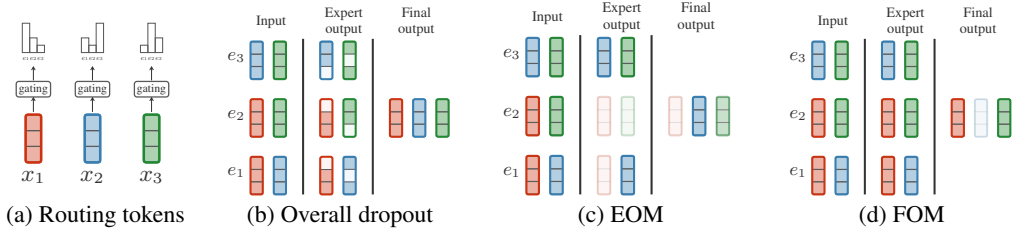
\begin{figure*}[!t]
\centering
\captionsetup[subfigure]{aboveskip=3pt,belowskip=3pt}
\begin{subfigure}[b]{0.16\textwidth}
\centering
\resizebox{.9\textwidth}{!}{
\begin{tikzpicture}
\def\halfw{3}
\def\yjump{1.0}
\def\xjump{0.8}
\newcommand{\plotdistrib}[5]{
    \def\x{#1}
    \def\y{#2}
    \def\pO{#3}
    \def\pB{#4}
    \def\pT{#5}
    \draw[dimgray] (\x, \y)  rectangle (\x+0.1, \y+\pO);
    \draw[dimgray] (\x+0.1, \y)  rectangle (\x+0.2, \y+\pB);
    \draw[dimgray] (\x+0.2, \y)  rectangle (\x+0.3, \y+\pT);
    \node[txt, scale=0.3] at (\x+0.05, \y-0.08) {$e_1$};
    \node[txt, scale=0.3] at (\x+0.15, \y-0.08) {$e_2$};
    \node[txt, scale=0.3] at (\x+0.25, \y-0.08) {$e_3$};
}
\newcommand{\plotvector}[4]{
\def\x{#1}
\def\y{#2}
\def\c{#3}
\def\o{#4}
\node[neurone, anchor=south,fill=\c!50, fill opacity=\o] (n1) at (\x*\xjump,\y*\yjump) {};
\node[neurone, anchor=south, above=-0.3pt of n1,fill=\c!50, fill opacity=\o] (n2) {};
\node[neurone, anchor=south, above=-0.3pt of n2,fill=\c!50, fill opacity=\o] (n3) {};
\draw[vector,\c] ([xshift=-\halfw pt, yshift=-0pt]n1.south)  rectangle ([xshift=\halfw pt, yshift=0pt]n3.north);
}

\plotvector{0}{0}{Set1A}{0.75}
\plotvector{1}{0}{Set1B}{0.75}
\plotvector{2}{0}{Set1C}{0.75}
\node[txt] at (0*\xjump, -0.2) {$x_1$};
\node[txt] at (1*\xjump, -0.2) {$x_2$};
\node[txt] at (2*\xjump, -0.2) {$x_3$};

\node[block, scale=0.5] (r1) at (0, \yjump) {gating};
\draw[conn, line width=0.2pt] (r1.south)+(0, -0.2) -- ++(0, 0);
\draw[conn, line width=0.2pt] (r1.north)+(0, 0.01) -- ++(0, 0.2);

\node[block, scale=0.5] (r2) at (\xjump, \yjump) {gating};
\draw[conn, line width=0.2pt] (r2.south)+(0, -0.2) -- ++(0, 0);
\draw[conn, line width=0.2pt] (r2.north)+(0, 0.01) -- ++(0, 0.2);

\node[block, scale=0.5] (r3) at (2*\xjump, \yjump) {gating};
\draw[conn, line width=0.2pt] (r3.south)+(0, -0.2) -- ++(0, 0);
\draw[conn, line width=0.2pt] (r3.north)+(0, 0.01) -- ++(0, 0.2);

\plotdistrib{-0.15}{1.5*\yjump}{0.4}{0.2}{0.1}
\plotdistrib{\xjump-0.15}{1.5*\yjump}{0.2}{0.1}{0.4}
\plotdistrib{2*\xjump-0.15}{1.5*\yjump}{0.1}{0.4}{0.2}

\end{tikzpicture}
}
\caption{Routing tokens}\label{fig:moe:routing}
\end{subfigure}\hspace{5pt}%
\begin{subfigure}[b]{0.22\textwidth}
\centering
\resizebox{.9\textwidth}{!}{
\begin{tikzpicture}
\def\drop{0.3}
\def\halfw{3}
\def\yjump{1.}
\def\xjump{0.4}

\newcommand{\plotvector}[3]{
\def\x{#1}
\def\y{#2}
\def\c{#3}
\node[neurone, anchor=south,fill=\c!50, fill opacity=0.75] (n1) at (\x*\xjump,\y*\yjump) {};
\node[neurone, anchor=south, above=-0.3pt of n1,fill=\c!50, fill opacity=0.75] (n2) {};
\node[neurone, anchor=south, above=-0.3pt of n2,fill=\c!50, fill opacity=0.75] (n3) {};
\draw[vector,\c] ([xshift=-\halfw pt, yshift=-0pt]n1.south)  rectangle ([xshift=\halfw pt, yshift=0pt]n3.north);
}

\newcommand{\plotvectordropO}[3]{
\def\x{#1}
\def\y{#2}
\def\c{#3}
\node[neurone, anchor=south,fill=\c!50, fill opacity=0.05, opacity=0.1] (n1) at (\x*\xjump,\y*\yjump) {};
\node[neurone, anchor=south, above=-0.3pt of n1,fill=\c!50, fill opacity=0.75] (n2) {};
\node[neurone, anchor=south, above=-0.3pt of n2,fill=\c!50, fill opacity=0.75] (n3) {};
\draw[vector,\c] ([xshift=-\halfw pt, yshift=-0pt]n1.south)  rectangle ([xshift=\halfw pt, yshift=0pt]n3.north);
}

\newcommand{\plotvectordropB}[3]{
\def\x{#1}
\def\y{#2}
\def\c{#3}
\node[neurone, anchor=south,fill=\c!50, fill opacity=0.75] (n1) at (\x*\xjump,\y*\yjump) {};
\node[neurone, anchor=south, above=-0.3pt of n1,fill=\c!50, fill opacity=0.05, opacity=0.1] (n2) {};
\node[neurone, anchor=south, above=-0.3pt of n2,fill=\c!50, fill opacity=0.75] (n3) {};
\draw[vector,\c] ([xshift=-\halfw pt, yshift=-0pt]n1.south)  rectangle ([xshift=\halfw pt, yshift=0pt]n3.north);
}

\newcommand{\plotvectordropT}[3]{
\def\x{#1}
\def\y{#2}
\def\c{#3}
\node[neurone, anchor=south,fill=\c!50, fill opacity=0.05, opacity=0.75] (n1) at (\x*\xjump,\y*\yjump) {};
\node[neurone, anchor=south, above=-0.3pt of n1,fill=\c!50, fill opacity=0.75] (n2) {};
\node[neurone, anchor=south, above=-0.3pt of n2,fill=\c!50, fill opacity=0.05, opacity=0.1] (n3) {};
\draw[vector,\c] ([xshift=-\halfw pt, yshift=-0pt]n1.south)  rectangle ([xshift=\halfw pt, yshift=0pt]n3.north);
}

\plotvector{0}{1}{Set1A}
\plotvector{1}{1}{Set1B}
\node[txt, rotate=0] at (-0.35, \yjump*1.38) {$e_1$};

\plotvector{0}{2}{Set1A}
\plotvector{1}{2}{Set1C}
\node[txt, rotate=0] at (-0.35, \yjump*2.38) {$e_2$};

\plotvector{0}{3}{Set1B}
\plotvector{1}{3}{Set1C}
\node[txt, rotate=0] at (-0.35, \yjump*3.38) {$e_3$};

\draw[conn, -](2*\xjump, 1*\yjump)-- (2*\xjump, 3.7*\yjump);

\plotvectordropT{3}{1}{Set1A}
\plotvectordropB{4}{1}{Set1B}

\plotvectordropT{3}{2}{Set1A}
\plotvectordropO{4}{2}{Set1C}

\plotvectordropO{3}{3}{Set1B}
\plotvectordropB{4}{3}{Set1C}

\draw[conn, -](5*\xjump, 1*\yjump)-- (5*\xjump, 3.7*\yjump);

\plotvector{6}{2}{Set1A}
\plotvector{7}{2}{Set1B}
\plotvector{8}{2}{Set1C}
\node[txt, font=\scriptsize] at (0.25, 4.0) {Input};
\node[txt, font=\scriptsize] at (1.5, 4.0) {Expert\\output};
\node[txt, font=\scriptsize] at (2.75, 4.0) {Final\\output};
\end{tikzpicture}
}
\caption{Overall dropout}\label{fig:moe:overall-dropout}
\end{subfigure}\hspace{5pt}%
\begin{subfigure}[b]{0.22\textwidth}
\centering
\resizebox{.9\textwidth}{!}{%
\begin{tikzpicture}
\def\drop{0.3}
\def\halfw{3}
\def\yjump{1.}
\def\xjump{0.4}

\newcommand{\plotvector}[3]{
\def\x{#1}
\def\y{#2}
\def\c{#3}
\node[neurone, anchor=south,fill=\c!50, fill opacity=0.75] (n1) at (\x*\xjump,\y*\yjump) {};
\node[neurone, anchor=south, above=-0.3pt of n1,fill=\c!50, fill opacity=0.75] (n2) {};
\node[neurone, anchor=south, above=-0.3pt of n2,fill=\c!50, fill opacity=0.75] (n3) {};
\draw[vector,\c] ([xshift=-\halfw pt, yshift=-0pt]n1.south)  rectangle ([xshift=\halfw pt, yshift=0pt]n3.north);
}

\newcommand{\plotdroppedvector}[3]{
\def\x{#1}
\def\y{#2}
\def\c{#3}
\node[neurone, anchor=south,fill=\c!50, fill opacity=0.1, opacity=0.1] (n1) at (\x*\xjump,\y*\yjump) {};
\node[neurone, anchor=south, above=-0.3pt of n1,fill=\c!50, fill opacity=0.1, opacity=0.1] (n2) {};
\node[neurone, anchor=south, above=-0.3pt of n2,fill=\c!50, fill opacity=0.1, opacity=0.1] (n3) {};
\draw[vector,\c, opacity=0.2] ([xshift=-\halfw pt, yshift=-0pt]n1.south)  rectangle ([xshift=\halfw pt, yshift=0pt]n3.north);
}

\newcommand{\plothalfwayvector}[3]{
\def\x{#1}
\def\y{#2}
\def\c{#3}
\node[neurone, anchor=south,fill=\c!50, fill opacity=0.5, opacity=0.5] (n1) at (\x*\xjump,\y*\yjump) {};
\node[neurone, anchor=south, above=-0.3pt of n1, fill=\c!50, fill opacity=0.5, opacity=0.5] (n2) {};
\node[neurone, anchor=south, above=-0.3pt of n2, fill=\c!50, fill opacity=0.5, opacity=0.5] (n3) {};
\draw[vector,\c, opacity=0.5] ([xshift=-\halfw pt, yshift=-0pt]n1.south)  rectangle ([xshift=\halfw pt, yshift=0pt]n3.north);
}

\plotvector{0}{1}{Set1A}
\plotvector{1}{1}{Set1B}
\node[txt, rotate=0] at (-0.35, \yjump*1.38) {$e_1$};

\plotvector{0}{2}{Set1A}
\plotvector{1}{2}{Set1C}
\node[txt, rotate=0] at (-0.35, \yjump*2.38) {$e_2$};

\plotvector{0}{3}{Set1B}
\plotvector{1}{3}{Set1C}
\node[txt, rotate=0] at (-0.35, \yjump*3.38) {$e_3$};

\draw[conn, -](2*\xjump, 1*\yjump)-- (2*\xjump, 3.7*\yjump);

\plotdroppedvector{3}{1}{Set1A}
\plotvector{4}{1}{Set1B}

\plotdroppedvector{3}{2}{Set1A}
\plotdroppedvector{4}{2}{Set1C}

\plotvector{3}{3}{Set1B}
\plotvector{4}{3}{Set1C}

\draw[conn, -](5*\xjump, 1*\yjump)-- (5*\xjump, 3.7*\yjump);

\plotdroppedvector{6}{2}{Set1A}
\plotvector{7}{2}{Set1B}
\plothalfwayvector{8}{2}{Set1C}
\node[txt, font=\scriptsize] at (0.25, 4.0) {Input};
\node[txt, font=\scriptsize] at (1.5, 4.0) {Expert\\output};
\node[txt, font=\scriptsize] at (2.75, 4.0) {Final\\output};

\end{tikzpicture}
}
\caption{EOM}\label{fig:moe:token-dropout}
\end{subfigure}\hspace{5pt}%
\begin{subfigure}[b]{0.22\textwidth}
\centering
\resizebox{.9\textwidth}{!}{
\begin{tikzpicture}
\def\drop{0.3}
\def\halfw{3}
\def\yjump{1.}
\def\xjump{0.4}

\newcommand{\plotvector}[3]{
\def\x{#1}
\def\y{#2}
\def\c{#3}
\node[neurone, anchor=south,fill=\c!50, fill opacity=0.75] (n1) at (\x*\xjump,\y*\yjump) {};
\node[neurone, anchor=south, above=-0.3pt of n1,fill=\c!50, fill opacity=0.75] (n2) {};
\node[neurone, anchor=south, above=-0.3pt of n2,fill=\c!50, fill opacity=0.75] (n3) {};
\draw[vector,\c] ([xshift=-\halfw pt, yshift=-0pt]n1.south)  rectangle ([xshift=\halfw pt, yshift=0pt]n3.north);
}

\newcommand{\plotdroppedvector}[3]{
\def\x{#1}
\def\y{#2}
\def\c{#3}
\node[neurone, anchor=south,fill=\c!50, fill opacity=0.1, opacity=0.1] (n1) at (\x*\xjump,\y*\yjump) {};
\node[neurone, anchor=south, above=-0.3pt of n1,fill=\c!50, fill opacity=0.1, opacity=0.1] (n2) {};
\node[neurone, anchor=south, above=-0.3pt of n2,fill=\c!50, fill opacity=0.1, opacity=0.1] (n3) {};
\draw[vector,\c, opacity=0.2] ([xshift=-\halfw pt, yshift=-0pt]n1.south)  rectangle ([xshift=\halfw pt, yshift=0pt]n3.north);
}

\newcommand{\plothalfwayvector}[3]{
\def\x{#1}
\def\y{#2}
\def\c{#3}
\node[neurone, anchor=south,fill=\c!50, fill opacity=0.5, opacity=0.5] (n1) at (\x*\xjump,\y*\yjump) {};
\node[neurone, anchor=south, above=-0.3pt of n1, fill=\c!50, fill opacity=0.5, opacity=0.5] (n2) {};
\node[neurone, anchor=south, above=-0.3pt of n2, fill=\c!50, fill opacity=0.5, opacity=0.5] (n3) {};
\draw[vector,\c, opacity=0.5] ([xshift=-\halfw pt, yshift=-0pt]n1.south)  rectangle ([xshift=\halfw pt, yshift=0pt]n3.north);
}

\plotvector{0}{1}{Set1A}
\plotvector{1}{1}{Set1B}
\node[txt, rotate=0] at (-0.35, \yjump*1.38) {$e_1$};

\plotvector{0}{2}{Set1A}
\plotvector{1}{2}{Set1C}
\node[txt, rotate=0] at (-0.35, \yjump*2.38) {$e_2$};

\plotvector{0}{3}{Set1B}
\plotvector{1}{3}{Set1C}
\node[txt, rotate=0] at (-0.35, \yjump*3.38) {$e_3$};

\draw[conn, -](2*\xjump, 1*\yjump)-- (2*\xjump, 3.7*\yjump);

\plotvector{3}{1}{Set1A}
\plotvector{4}{1}{Set1B}

\plotvector{3}{2}{Set1A}
\plotvector{4}{2}{Set1C}

\plotvector{3}{3}{Set1B}
\plotvector{4}{3}{Set1C}
\draw[conn, -](5*\xjump, 1*\yjump)-- (5*\xjump, 3.7*\yjump);

\plotvector{6}{2}{Set1A}
\plotdroppedvector{7}{2}{Set1B}
\plotvector{8}{2}{Set1C}
\node[txt, font=\scriptsize] at (0.25, 4.0) {Input};
\node[txt, font=\scriptsize] at (1.5, 4.0) {Expert\\output};
\node[txt, font=\scriptsize] at (2.75, 4.0) {Final\\output};

\end{tikzpicture}
}
\caption{FOM}\label{fig:moe:fom}
\end{subfigure}
\vspace{-5pt}
\caption{Illustration of Expert/Final Output Masking (EOM/FOM) in contrast to overall dropout for MoE layers: a color represents a token, and each token is dispatched to two experts. Faded colors correspond to dropped units or masked outputs. Note that EOM and FOM are always combined with overall dropout.
}
\end{figure*}

The motivation behind MoE models is to allow different parameters to model different aspects of the input space. The added expert capacity should help higher resource language pairs that might otherwise be constrained to share the same capacity with many other language pairs. Besides, increasing model capacity should reduce interference, thus benefiting tasks of all resource levels.

Although overall dropout is sufficient to regularize dense models, it is not enough for MoE models (see \Cref{fig:ablation:ppl}). 
To address the issue of over-fitting of MoE models on low-resource tasks, we propose a series of architectural changes that improve the performance on low-resource language pairs with MoE models in \Cref{sec:eom,sec:fom,sec:cmr}. 
In \Cref{sec:cl}, we devise and study a simple but effective curriculum learning strategy as another approach to reduce the over-fitting on low-resource directions.

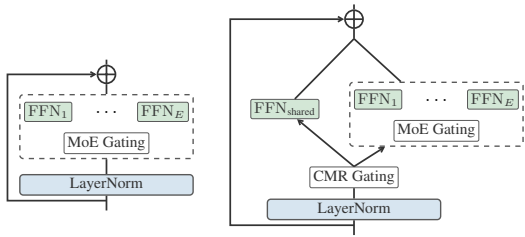
\begin{figure}[t]
    \centering
    \resizebox{.9\columnwidth}{!}{%
    \begin{tikzpicture}
\begin{scope}
    \node (input) at (0, 0) {};
    \node[layer, above=3mm of input, fill=Set1B!20] (moe_norm) {LayerNorm};
    \node[layer, above=3mm of moe_norm, minimum height=1.3cm, dashed]  (moe) {};
    \node[block, anchor=south] at ([yshift=3pt]moe.south) (router) {MoE Gating};
    \node[block, anchor=west, fill=Set1C!20] at ([yshift=3mm, xshift=1mm]moe.west) (ffn1) {$\ffn 1$};
    \node[anchor=south, above =23pt of moe.south] (dots) {$\ldots$};
    \node[block, anchor=east, fill=Set1C!20] at ([yshift=3mm, xshift=-1mm]moe.east) (ffn1) {$\ffn E$};
    \node[residual, above=2mm of moe] (res2) {};
    \draw[conn] ([yshift=-1mm]moe_norm.south) -- (
        [xshift=-2cm, yshift=-1mm]moe_norm.south) |- (res2.west);

    \draw[conn, -] (input) -- (moe_norm) -- (moe) -- (res2) -- + (0, 0.3);
\end{scope}

\begin{scope}[xshift=5cm, yshift=-5mm]
    \node (input) at (0, 0) {};
    \node[layer, above=3mm of input, fill=Set1B!20] (moe_norm) {LayerNorm};
    \node[block, anchor=south] at ([yshift=6pt]moe_norm.north) (clsr_router) {CMR Gating};
    \node[layer, above right=4mm and -10mm of clsr_router, minimum height=1.3cm, dashed]  (moe) {};
    \node[block, fill=Set1C!20, above left=-1mm and 6mm of moe.west] (shared_ffn) {$\ffn{\text{shared}}$};
    \node[block, anchor=south] at ([yshift=3pt]moe.south) (router) {MoE Gating};
    \node[block, anchor=west, fill=Set1C!20] at ([yshift=3mm, xshift=1mm]moe.west) (ffn1) {$\ffn 1$};
    \node[anchor=south, above =23pt of moe.south] (dots) {$\ldots$};
    \node[block, anchor=east, fill=Set1C!20] at ([yshift=3mm, xshift=-1mm]moe.east) (ffn1) {$\ffn E$};
    \node[residual, above=28mm of clsr_router] (res2) {};
    \draw[conn] ([yshift=-1mm]moe_norm.south) -- (
        [xshift=-2.5cm, yshift=-1mm]moe_norm.south) |- (res2.west);

    \draw[conn, -] (input) -- (moe_norm) -- (clsr_router);
    \draw[conn] (clsr_router.north) -- (shared_ffn);
    \draw[conn, -] (shared_ffn) -- ([yshift=2.6cm]clsr_router.north);
    \draw[conn] (clsr_router.north) -- (moe);
    \draw[conn, -] (moe) -- ([yshift=2.6cm]clsr_router.north);
    \draw[conn, -] ([yshift=2.6cm]clsr_router.north) -- (res2) -- +(0, 0.3);

\end{scope}
\end{tikzpicture}
    }
    \vspace{-5pt}
    \caption{Illustration of \clsr{} (\clsrabbrv{}) showing a residual block in a Transformer layer with regular MoE (left) vs. \clsrabbrv{} (right).}
    \label{fig:modeling:clsr}
\end{figure}

\subsection{\moetokendropout{} (\moetokendropoutabbrv{}).}\label{sec:eom}
In this proposed regularization strategy, we mask the \textit{expert output} for a random fraction ($\moetokdrop$) of the input tokens. For input tokens with dropped expert outputs, the first and/or second expert is effectively skipped, as illustrated in \Cref{fig:moe:token-dropout}.
Note that although this masking will zero out some combination weights $\mathcal G_{t,e}$ in \Cref{eqn:moe-layer}, it will not affect the weights used in the load balancing loss.

\subsection{Final Output Masking (\textsc{FOM}).}\label{sec:fom}
A simpler alternative to \moetokendropoutabbrv{} would be to mask the combined expert output for a random fraction of tokens, i.e., the last stage in \Cref{fig:moe:fom}. We denote with $p_\text{fom}$ the fraction of tokens masked with this regularization method. Note that this type of masking is more generic as it can be applied to dense models as well.

\subsection{\clsr{} (\clsrabbrv{}).}\label{sec:cmr}
Instead of randomly dropping a proportion of activations or masking expert outputs, we consider the option of letting the model learn which tokens need the extra capacity or specialization of MoE layers, and which tokens are better routed to a limited-capacity shared layer.
Inspired by~\citet{zhang2021clsr}'s CLSR-Gate, we design \clsr{} layers (\clsrabbrv{} for short). As depicted in \Cref{fig:modeling:clsr}, we augment MoE layers with a binary gate that determines the weights associated with two branches of the computational graph: \textbf{(1)} a shared dense FFN sublayer ($\ffn{\text{shared}}$) and \textbf{(2)} an MoE layer with its own $E$ expert FFN sublayers. For an input token $x_t$, the output of \clsrabbrv{} is evaluated as follows:
\begin{align}
g(x_t) & = \operatorname{sigmoid}(W_{\clsrabbrv{}}\cdot x_t),\\
\operatorname{\clsrabbrv{}}(x_t) & = (1 - g(x_t))\cdot \ffn{\text{shared}}(x_t) \\
&+ g(x_t) \cdot \moe(x_t),
\label{eqn:modeling:clsr}
\end{align}\vspace{-0.5pt}
where $W_{\clsrabbrv{}}$ are the weights of the \clsrabbrv{}'s binary gate.
$W_{\clsrabbrv{}}$ is trained by optimizing translation accuracy under a budget constraint $\clsrbudget$. For a mini-batch with $T$ tokens, this amounts to adding the following auxiliary loss term ($L_{\clsrabbrv{}}$) to the loss function in \cref{eqn:moeloss}:
\begin{align}
& L_{\clsrabbrv{}} = \frac1T\cdot \sum\limits_{t=1}^T \left|g(x_t)  - \clsrbudget \right|,    
\label{eqn:modeling:clsr:loss}\\
& L = L_{\text{MT}} + \lambda_{\text{MoE}} L_{\text{MoE}} + \lambda_{\clsrabbrv{}} L_{\clsrabbrv{}}.
\end{align}
We use the budget parameter $\clsrbudget$ to limit the effective capacity of MoE layers, thus providing a regularizing effect; at $\clsrbudget{=}0$, the model is dense, practically pushing all tokens through $\ffn{\text{shared}}$, and at $\clsrbudget{=}1$, the model is free to always route tokens through the high-capacity MoE layer. 

To reduce over-fitting, we experiment with zeroing out a fraction of the \clsrabbrv{} gates $g(x_t)$ in the mini-batch; we denote this fraction with $\clsrdrop{}$. %
This means that we force $\clsrdrop{}\%$ tokens in the mini-batch to only take the route of $\ffn{\text{shared}}$.

\subsection{Curriculum Learning}\label{sec:cl}
We next explore alternative methods of regularization by means of Curriculum Learning (CL). 
We propose to start training with high-resource pairs first, then introduce low-resource pairs, prone to over-fit, later in phases. 
To derive the phases of the curriculum, we first train a vanilla MoE model (without CL), then we partition the tasks (translation directions) into $n$ bins
$\{b_1,\ldots, b_n\}$. 
If $\clupdates$ is the total number of training updates, we introduce each bin $b_i$ after $\clupdates - k_i$ updates.
We compare two partitioning strategies for when and what directions to add at every phase.
\begin{enumerate}
\itemsep0pt
\item \textit{Count-based}: we empirically partition based on training example counts.
\item \textit{Step-based}: partition based on the step where we observed a task to start over-fitting. See \Cref{alg:cl:step}.
\end{enumerate}

\begin{algorithm}[!t]
\small
\caption{Partitioning for \textit{step-based} CL 
}\label{alg:cl:step}
\begin{algorithmic}[1]
\State\algorithmicrequire~number of bins $n$, a set of tasks $\mathcal T$, the maximum number of updates $\clupdates$, the step corresponding to the best validation perplexity $s_{\text{best}}: \mathcal T \to [0, \clupdates]$.
\LineComment  For $s_{\text{best}}$, we take the max if multiple
\State\algorithmicensure~Partitioning of $\mathcal T$ into $n$ bins $\mathbf b{=}(b_1, \ldots, b_n)$, characteristic step for each bin $\mathbf k{=}(k_1, \ldots, k_n)$.
\LineComment The bin $b_i$ will be introduced at $\clupdates - k_i$. 
    \State $s_\text{max} = \max_{t\in\mathcal T} s_\text{best}(t)$, $s_\text{min} = \min_{t\in\mathcal T} s_\text{best}(t)$
    \State $\Delta = \dfrac{s_\text{max} - s_\text{min}}{n-1}$.
    \For{$i \in \{1\ldots n\}$}
    \State $b_i = \varnothing, \; k_i=s_\text{max} - (i-1)\Delta$.
    \EndFor
    \For{$t \in \mathcal T$}
    \State $c_t = \argmin_{1\leq i \leq n} |s_\text{best}(t) - k_i|$ 
    \State $b_{c_t} = b_{c_t} \cup \{t\}$
    \LineComment assign to the closest bin wrt. its characteristic step.
    \EndFor
\end{algorithmic}
\end{algorithm}

\section{Experimental Setup}\label{sec:setup}
\subsection{\ablation}
We construct a multilingual machine translation benchmark consisting of 53 languages and a total of 110 translation directions.
Our \ablation~consists of 45 directions out of English (aggregated as \langcode{eng-xx}), 45 directions into English (aggregated as \langcode{xx-eng}) and 20 non-English directions (aggregated as \langcode{xx-yy}).
In terms of resource level, there are 40 high-resource and 70 low-resource directions, out of which 22 are very low-resource.\footnote{
    We follow the categorization in \citet{nllb2022}; a language is low-resource if there are fewer than 1M publicly available, de-duplicated bitext samples with any other language, very low-resource if fewer than 100K.
}
The training data is composed of publicly available bitext in all 110 language directions (primary data in \citet{nllb2022})
and large-scale mined data \citep{heffernan2022bitext, nllb2022} in English-centric directions.
There are a total of 2$\times$847M examples in this benchmark. 
For a detailed listing of the directions, see \Cref{appendix:data}.

\paragraph{Segmentation with SentencePiece.}
To tokenize our text sequences, we train a single SentencePiece (SPM)~\citep{kudo2018sentencepiece} model for all languages.\footnote{202 languages in total, including the ones not part of our \ablation.}
The vocabulary size of our trained SPM model is 256,000. For more on this SPM model, see \citet{nllb2022}.

\section{Results}\label{sec:results}
\begin{table}[!t]
\centering
\resizebox{.95\columnwidth}{!}{
    \begin{tabular}{lcHHccHHcc}
    \toprule
     & \multicolumn{4}{c}{\langcode{eng-xx}} &  \multicolumn{4}{c}{\langcode{xx-eng}}  &  \multicolumn{1}{c}{\langcode{xx-yy}} \\
      \cmidrule(l){2-5}
      \cmidrule(lr){6-9}
      \cmidrule(r){10-10}
        ~ & all & high & low & v.low & all & high & low & v.low & all \\
        \midrule
\textsc{Dense \modelsizesm{}} & & & & & & & & & \\
\phantom{ab}Dense & 41.7	& 54.2 &	36.6 &	30.4 &	51.1 &	61.5 &	46.8 &	44.0 &	39.4 \\
\phantom{ab}MoE-64 &	43.0 &	55.3 &	37.9 &	30.3 &	52.6 &	63.3 &	48.3 &	44.2 &	39.8 \\
\cmidrule(l){2-10}
\phantom{ab}Dense ($\drop{=}0.1$) &	41.9 &	54.1 &	37.0 &	31.1 &	51.8 &	61.8 &	47.8 &	45.3 &	39.6 \\
\phantom{ab}MoE-64 ($\drop{=}0.1$) &  \bf 43.6 &	 \bf 55.8 &	 \bf 38.7 &	 \bf 32.0 &  \bf 53.4 &	 \bf 63.6 &	 \bf 49.3 &	 \bf 45.9 &	 \bf 41.1 \\
\midrule
\midrule
\textsc{Dense \modelsize{}} & & & & & & & & & \\
\phantom{ab}Dense & 43.3 &	55.4 &	38.4 &	31.6 &	53.5 &	63.6 &	49.4 &	46.5 &	41.3 \\
\phantom{ab}MoE-64 &	43.3 &	 55.9 &	38.2 &	29.7 &	52.9 &	 \bf 63.9 &	48.4 &	43.7 &	39.3 \\
\cmidrule(l){2-10}
\phantom{ab}Dense ($\drop{=}0.1$)$\dagger$	 &  43.7	&  55.4 &	39.0 &	  \bf 33.1 &	\bf 54.4 &	63.8 &	  \bf 50.6 & \bf 47.9 &	  \bf 41.9 \\
\phantom{ab}MoE-64 ($\drop{=}0.3$)$\dagger$ & \bf 44.3 &	 \bf 56.0 &	  \bf 39.5 &	32.5 &	  \bf  54.4 &	 \bf 63.9 &	  \bf 50.6 &	47.7 &	\bf  41.9 \\
    \bottomrule
    \end{tabular}}
    \caption{Validation set \chrf{} of vanilla MoE with and without overall dropout.
     $\dagger$ indicates best of sweep.
    }
    \label{tbl:modeling:moe}
\end{table}

\begin{figure*}[!t]
\centering
\begin{tikzpicture}[scale=1]
\def\lrpair{eng-kon}
\def\hrpair{eng-fra}
\def\yminlr{10}
\def\ymaxlr{60}
\def\yminhr{1.9}
\def\ymaxhr{2.19}

\begin{groupplot}[
        group style={group size=3 by 2, horizontal sep=20pt, vertical sep=20pt},
        grid=both,
        minor tick num=1,
        label style={font=\scriptsize},
        tick label style={font=\tiny},
        y tick label style={font=\tiny, xshift=2pt},
        x tick label style={font=\tiny, yshift=1pt},
        scaled x ticks=false,
        y axis line style=-,
        cycle list name=CustomListWithMarkers,
        height=3.4cm,
        width=5.2cm,
        xtick={10000,40000,...,100000},
        xticklabels={10k,40k,70k,100k},
        xmin=10000,xmax=100000,
       ]
\nextgroupplot[
    ymin=\yminlr,ymax=\ymaxlr,
    legend style={at={(0.01,0.99)},anchor=north west, font=\scriptsize, nodes={scale=1, transform shape}},
    ylabel={Perplexity},
   ]
    \foreach \model/\modellabel in {%
        dense_drop0/Dense($\drop{=}0$),%
        dense_drop0.3/Dense($\drop{=}0.3$)%
        }{
        \addplot+
        table [y=\lrpair,x=updates]{results/ablation/\model.dat};
        \addlegendentryexpanded{\modellabel}
    }

\nextgroupplot[
    ymin=\yminlr,ymax=\ymaxlr,
    legend style={at={(0.01,0.99)},anchor=north west, font=\scriptsize, nodes={scale=1.0, transform shape}},
       ]
    \pgfplotsset{cycle list shift=2}
    \foreach \model/\modellabel in {%
        moe64_drop0/MoE64($\drop{=}0$),%
        moe64_drop0.3/MoE64($\drop{=}0.3$)%
        }{
        \addplot+
        table [y=\lrpair,x=updates]{results/ablation/\model.dat};
        \addlegendentryexpanded{\modellabel}
    }
    \legend{}
\nextgroupplot[
    ymin=\yminlr,ymax=\ymaxlr,
    legend style={at={(0.01,0.99)},anchor=north west, font=\scriptsize, nodes={scale=0.9, transform shape}},
   ]
    \pgfplotsset{cycle list shift=3}
    \foreach \model/\modellabel in {%
        moe64_drop0.3/MoE64($\drop{=}0.3$),%
        moe64_drop0.3_tok0.2/MoE64($\drop{=}0.3$\,$\moetokdrop{=}0.2$)%
        }{
        \addplot+
        table [y=\lrpair,x=updates]{results/ablation/\model.dat};
        \addlegendentryexpanded{\modellabel}
    }
\nextgroupplot[
        ymin=\yminhr,ymax=\ymaxhr,
        xlabel=updates,
        legend style={at={(0.99,0.99)},anchor=north east, font=\tiny, nodes={scale=0.7, transform shape}},
        ylabel=Perplexity,
       ]
    \foreach \model/\modellabel in {%
        dense_drop0/Dense($\drop{=}0$),%
        dense_drop0.3/Dense($\drop{=}0.3$)%
        }{
        \addplot+
        table [y=\hrpair,x=updates]{results/ablation/\model.dat};
        \addlegendentryexpanded{\modellabel}
    }
    \legend{}
\nextgroupplot[
        ymin=\yminhr,ymax=\ymaxhr,
        xlabel=updates,
        legend style={at={(0.99,0.99)},anchor=north east, font=\scriptsize, nodes={scale=1.0, transform shape}},
       ]
    \pgfplotsset{cycle list shift=2}
    \foreach \model/\modellabel in {%
        moe64_drop0/MoE64($\drop{=}0$),%
        moe64_drop0.3/MoE64($\drop{=}0.3$)%
        }{
        \addplot+
        table [y=\hrpair,x=updates]{results/ablation/\model.dat};
        \addlegendentryexpanded{\modellabel}
    }
\nextgroupplot[
    ymin=\yminhr,ymax=\ymaxhr,
    xlabel=updates,
    legend style={at={(0.99,0.99)},anchor=north east, font=\tiny, nodes={scale=0.7, transform shape}},
   ]
    \pgfplotsset{cycle list shift=3}
    \foreach \model/\modellabel in {%
        moe64_drop0.3/MoE64($\drop{=}0.3$),%
        moe64_drop0.3_tok0.2/MoE64($\drop{=}0.3$\,$\moetokdrop{=}0.2$)%
        }{
        \addplot+
        table [y=\hrpair,x=updates]{results/ablation/\model.dat};
        \addlegendentryexpanded{\modellabel}
    }
    \legend{}
\end{groupplot}
\node[txt, anchor=south, rotate=90, font=\small]  at (rel axis cs: -0.22, 0.5) {\langcode{eng-kon}};
\node[txt, anchor=south, rotate=90, font=\small]  at (rel axis cs: -0.22, -0.85) {\langcode{eng-fra}};

\node[txt, anchor=south, align=center, font=\small]  at (rel axis cs: 0.5, 0.95) {Dense + dropout};
\node[txt, anchor=south, align=center, font=\small]  at (rel axis cs: 1.7, 0.95) {MoE + dropout};

\node[txt, anchor=south, align=center, font=\small]  at (rel axis cs: 2.9, 0.95) {MoE + EOM};

\end{tikzpicture}
\vspace{-5pt}
\caption{Validation perplexities with Various dropout strategies for a low-resource direction (\langcode{eng-kon} in the top row) and a high-resource direction (\langcode{eng-fra} in the bottom row). 
}\label{fig:ablation:ppl}
\end{figure*}
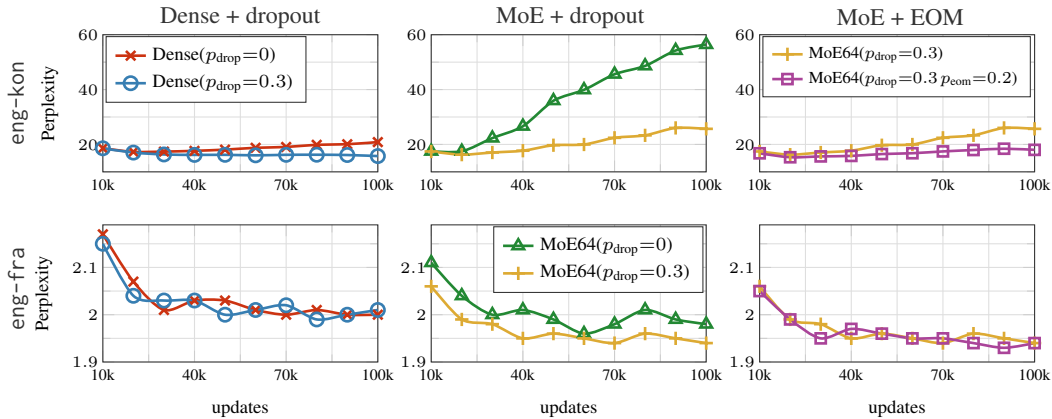
All MoE sub-layers have $E{=}$\moeExperts{} experts\footnote{$E=64$ is close to the number of languages in the benchmark, i.e., 53}. All models are trained for 100k updates with an effective batch size of 1M tokens per update.
See \Cref{appendix:training_hparams} for additional details.
We use the \chrf{} metric~\citep{popovic:2017:WMT} to compare the model performance.
We report averages in each set of directions: \langcode{eng-xx}, \langcode{xx-eng} and \langcode{xx-yy} as \textit{all}. For \langcode{eng-xx} and \langcode{xx-eng}, and when relevant, we breakdown the pairs by resource level: high-resource (\textit{high}), low-resource (\textit{low}) and very low resource (\textit{v.low}).
\subsection{Vanilla (un-regularized) MoE}
When looking at un-regularized models (without overall dropout), we see in \Cref{tbl:modeling:moe}, that when the backbone dense model has \modelsizesm{} parameters, the MoE model, while computationally similar, shows +1.3, +1.5 and +0.4 \chrf{} improvements on \langcode{eng-xx}, \langcode{xx-eng} and \langcode{xx-yy} respectively. When focusing on the very low resource pairs (v.low), the performance actually drops on \langcode{eng-xx} (-0.1 \chrf{}) signaling an over-fitting issue.
When scaling the backbone to 1.3B, we see even more over-fitting on \textit{v.low} directions (-1.9\chrf{} in \langcode{eng-xx} and -2.8 \chrf{} in \langcode{xx-eng}).

Adding overall dropout\footnote{sweeping over $\drop{\in}\{0.1, 0.2, 0.3\}$} significantly improves the performance of MoE models in both the \modelsizesm{} and \modelsize{} variants.
Importantly, when increasing the dropout to 0.1 for the small MoE (\modelsizesm{}), we see that the relative decline of -0.1 \chrf{}, turns into an improvement of +0.9 \chrf{} for \langcode{eng-xx} v.low pairs.
Once we scale the computational cost per update (\modelsize{}), tuned overall dropout does not fix the over-fitting of very low-resource pairs.

In \Cref{fig:ablation:ppl}, we observe in the case of \langcode{eng-kon}, a very low-resource pair, that the model continues to face significant over-fitting when trained for 100k updates. This is unsurprising, as iterating over a small training set with large capacity causes over-fitting. Training for more steps is important for high-resource pairs, but we want to avoid negatively affecting low-resource pairs in the process. 

\subsection{Regularizing MoEs}
\begin{table*}[!ht]
\centering
\resizebox{.98\linewidth}{!}{
    \begin{tabular}{lccccccccc}
    \toprule
     & \multicolumn{4}{c}{\langcode{eng-xx}} &  \multicolumn{4}{c}{\langcode{xx-eng}}  &  \multicolumn{1}{c}{\langcode{xx-yy}} \\
      \cmidrule(l){2-5} 
      \cmidrule(lr){6-9}
      \cmidrule(r){10-10}
        ~ & all & high & low & v.low & 
        all & high & low & v.low &
        all \\ 
        \midrule
Baseline $\drop{=}0.3$ $\dagger$ & 44.3	&	\bf 56.0	&	39.5	&	32.5	&	54.4	&	63.9	&	50.6	&	47.7	&	41.9 \\
\midrule
\moetokendropoutabbrv{} ($\drop{=}0.3$, $\moetokdrop{=}0.1$)$\dagger$	& \bf 44.7	&	55.9	&	\bf 40.1	& \bf	33.4	&	54.8	&	\bf 64.3	&	51.0	&	48.3	&	\bf 42.5 \\
FOM  ($\drop{=}0.2$, $p_\text{fom}{=}0.3$) $\dagger$	& 44.4	&	55.7	&	39.8	&	33.1	&	\bf 55.0	&	\bf 64.3	&	\bf 51.3	&	\bf 48.8	&	\bf 42.5 \\

\midrule

\localgatedropout{}~($\drop{=}0.3$, $p_{gd}{=}0.2$)~\citep{liu2022gating} $\dagger$ & 44.4 &	55.7 & 39.8 & 	33.0 &	54.8 &	64.1 &	51.0 &	48.5 &	42.3 \\

\midrule
\clsrabbrv{} top-1  ($\drop{=}0.3$, $\clsrdrop{=}0.1$, $\clsrbudget{=}0.6$) $\dagger$	& 44.2	&	55.8	&	39.5	&	33.2	&	54.9	&	\bf 64.3	&	51.1	& 48.7	&	42.3 \\

\midrule
\midrule
\clsrabbrv{} top-2 ($\drop{=}0.3$, $\clsrdrop{=}0.2$) $\dagger$ & \bf 46.2	&	\bf 56.2	&\bf	41.8	&	\bf 35.7	& \bf	55.1	&	\bf 64.7	&	\bf 51.5	&	\bf 49.2	&	\bf 42.8 \\
    \bottomrule
    \end{tabular}}
    \caption{Comparison of Various Regularization Strategies applied to an MoE-64 baseline. In each column, we bold the best results out of the first six rows (computationally comparable), and we bold results from the last row (\clsrabbrv{} top-2) if they outperform the other models. $\dagger$ signals that this model is best of sweep.
    }
    \label{tbl:modeling:moe:regul}
\end{table*}

For the rest of this paper, we use the 1.3B variant as our backbone, to which we add MoE layers with $E{=}64$ experts.

\paragraph{Experimental Setup.}
We consider the MoE model with an overall dropout rate of 0.3 ($\drop{=}0.3$), best performing after a sweep of $\drop\in\{0.1, 0.2, 0.3\}$ to be our baseline.\footnote{
Initial experiments separating the dropout rates of shared and MoE blocks showed that the best values align.
}

\noindent
In each of the sweeps below, we choose the best variant based on the average \chrf{} score on the validation set. 

\noindent
For \moetokendropoutabbrv{} and \fom{}, we sweep over the values of $(\drop, p_\text{eom/fom})\in \{0.1, 0.2, 0.3\}^2$.

\noindent
For \clsrabbrv{}, and in order to keep the compute equivalent to the baseline MoE, we use top-1 instead of the top-2 gating used in previous experiments. We fix $\drop{=}0.3$ and sweep over the \clsrabbrv{} parameters ($\clsrdrop{}$, $\clsrbudget{}$).
We also train a \clsrabbrv{} top-2 model, although not compute-equivalent to the baseline MoE, it provides insight into performance under a large compute budget. For \clsrabbrv{} top-2, 
we fix $\drop{=}0.3$
and sweep over the values of 
$\clsrdrop{}\in\{0.1,0.2,0.3\}$.
We set $\lambda_\text{CMR}$ to 0.1 in all our CMR experiments.

\noindent
We additionally compare our methods to \localgatedropout{}~\citep{liu2022gating}, a method in which we route tokens with probability $p_{gd}$ to the local experts, thus skipping the \texttt{All-to-All} communication between GPUs.
We sweep over the values of $(\drop, p_{gd})\in \{0.1, 0.2, 0.3\}^2$.

For each model, we report \chrf{} averages on the validation set in 3 groups of directions: \langcode{eng-xx}, \langcode{xx-eng} and \langcode{xx-yy}, broken down \wrt to resource levels: high, low and very low (v.low) for \langcode{eng-xx} and \langcode{xx-eng}.

\begin{table}[!t]
\centering
\resizebox{.95\columnwidth}{!}{
    \begin{tabular}{lcHHccHHcc}
    \toprule
     & \multicolumn{4}{c}{\langcode{eng-xx}} &  \multicolumn{4}{c}{\langcode{xx-eng}}  &  \multicolumn{1}{c}{\langcode{xx-yy}} \\
      \cmidrule(lr){2-5} 
      \cmidrule(lr){6-9}
      \cmidrule(lr){10-10}
        ~ & all & high & low & v.low & 
        all & high & low & v.low &
        all \\ 
\midrule
top-1 $\clsrbudget{=}0.4$	& \bf 44.2	&	55.8	&	39.4	&	32.8	&	54.4	&	63.9	&	50.5	&	48.0	&	42.1 \\
\phantom{top-1}+ $\clsrdrop{=}0.1$	& 43.9	&	55.7	&	39.1	&	\bf 33.0	&	\bf 54.9	&	64.3	&	\bf 51.1	&	\bf 48.6	&	\bf 42.3 \\
\midrule
top-1, $\clsrbudget{=}0.8$	& 44.5	&	\bf 56.2	&	39.7	&	32.9	&	\bf 54.3	&	64.2	&	50.2	&	47.4	&	\bf 42.2 \\
\phantom{top-1}+ $\clsrdrop{=}0.1$	& \bf 44.6	&	56.0	&	\bf 40.0	&	\bf 33.5	&	\bf 54.3	&	\bf 64.4	&	50.2	&	\bf 47.7	&	\bf 42.2 \\
\midrule
\midrule
top-1 $\clsrdrop{=}0.1$ \\
\phantom{top-1} + $\clsrbudget{=}0.2$	& 43.8	&	55.7	&	39.0	&	32.7	&	54.5	&	63.5	&	50.8	&	48.5	&	42.2 \\
\phantom{top-1} + $\clsrbudget{=}0.4$	& 43.9	&	55.7	&	39.1	&	33.0	&	\bf 54.9	&	64.3	&	\bf 51.1	&	48.6	&	\bf 42.3 \\
\phantom{top-1} + $\clsrbudget{=}0.6$	& 44.2	&	55.8	&	39.5	&	33.2	&	\bf 54.9	&	64.3	&	\bf 51.1	&	\bf 48.7	&	\bf 42.3 \\
\phantom{top-1} + $\clsrbudget{=}0.8$ 	& \bf 44.6	&	\bf 56.0	&	\bf 40.0	&	\bf 33.5	&	54.3	&	\bf 64.4	&	50.2	&	47.7	&	42.2 \\
\midrule
\midrule
top-2 $\clsrbudget{=}0.8$ & 44.6	&	56.0	&	39.9	&	33.1	&	54.3	&	64.0	&	50.3	&	47.2	&	41.9 \\
\phantom{top-1}+ $\clsrdrop{=}0.2$  & \bf 46.2	&	\bf 56.2	&\bf	41.8	&	\bf 35.7	& \bf	55.1	&	\bf 64.7	&	\bf 51.5	&	\bf 49.2	&	\bf 42.8 \\

    \bottomrule
    \end{tabular}}
    \caption{
    Sweep over hyperparameters for MoE-64 \clsrabbrv{}: The budget $\clsrbudget$, the CMR gate dropout $\clsrdrop$. 
    We bold the best results in each column.
    }
    \label{tbl:modeling:moe:cmr}
\end{table}

\paragraph{Results.}

In terms of alleviating the over-fitting issue, the last column of \Cref{fig:ablation:ppl} shows that \moetokendropoutabbrv{} leads to better regularization and less over-fitting on low-resource tasks compared to overall dropout. 
In terms of translation quality, we observe in~\Cref{tbl:modeling:moe:regul} gains of +0.4 \chrf{} across all pairs into English and +0.6 \chrf{} across non-English pairs for MoE \moetokendropoutabbrv{} compared to the MoE baseline.
For out of English, the largest gains are observed on low and very low-resource languages; +0.6 and 0.9 \chrf{} respectively.

With FOM, we see in \Cref{tbl:modeling:moe:regul} gains over the baseline MoE of 
+0.1 \chrf{} across \langcode{eng-xx} pairs,
+0.6 \chrf{} across \langcode{xx-eng} pairs
and +0.6 \chrf{} across \langcode{xx-yy} pairs.
For into English, the largest gains are observed on low and very low-resource languages; +0.7 and 1.1 \chrf{}.
Compared to the best \moetokendropoutabbrv{} model, FOM under-performs slightly on \langcode{eng-xx} (-0.3 \chrf{}) but outperforms on \langcode{xx-eng} (+0.2 \chrf{}); when averaging over all pairs, the two models achieve the same \chrf{} score of 48.4.

We look in \Cref{tbl:modeling:moe:cmr} at the impact of the budget $\clsrbudget$ and the dropout $\clsrdrop$.
We observe that $\clsrdrop$ is a necessary ingredient in \clsrabbrv{} top-2; in the last two rows of \Cref{tbl:modeling:moe:cmr}, adding $\clsrdrop$ improves the performance across the board, particularly in \langcode{en-xx} and \langcode{xx-en} very low directions (+2.6 and +2.0 \chrf{}, respectively). 
With top-1, $\clsrdrop$ is less critical as it barely affects the overall performance, but does help on \langcode{eng-xx} and \langcode{xx-eng} very low pairs.
In the middle section of \Cref{tbl:modeling:moe:cmr}, we note that \clsrabbrv{} top-1 is not sensitive to the exact value of $\clsrbudget$, but, at low budget $\clsrbudget$ (less capacity), model performance significantly drops on \langcode{eng-xx} across all pairs. 
Pairs in \langcode{xx-eng}, on the other hand, favor a mid-range budget value.

In \Cref{tbl:modeling:moe:regul} for \clsrabbrv{} top-1, we see +0.4 \chrf{} across all pairs into English, and +0.4 \chrf{} across non-English pairs. 
Improvements are larger for out of English low and very low-resource languages, with +0.5 and +1.0 \chrf{} respectively.
For \clsrabbrv{} top-2, we see +1.9 \chrf{} across all pairs out of English and +0.9 \chrf{} across non-English pairs.
The improvements are largest for low and very low-resource languages, with +2.3 and +3.2 \chrf{} out of English, and +0.9 and +1.5 into English. \clsrabbrv{} top-2 is computationally more expensive by 23\% because of the additional shared FFN layer at the level of each MoE layer in the model. 

We find that \localgatedropout{} performs better than the baseline MoE, but is outperformed by all of our proposed methods. Overall, these results demonstrate that \moetokendropoutabbrv{}, \fom{}, and \clsrabbrv{} strategies help improve on top of vanilla MoE. 

\subsection{CL}
\begin{table}[!t]
\centering
\resizebox{\columnwidth}{!}{
    \begin{tabular}{lcHHccHHcc}
    \toprule
     & \multicolumn{4}{c}{\langcode{eng-xx}} &  \multicolumn{4}{c}{\langcode{xx-eng}}  &  \multicolumn{1}{c}{\langcode{xx-yy}} \\
      \cmidrule(lr){2-5} 
      \cmidrule(lr){6-9}
      \cmidrule(lr){10-10}
        ~ & all & high & low & v.low & all & high & low & v.low & all \\ 
        \midrule
MoE-64 &	44.3 &	 \bf 56.0 &	39.5 &	32.5 &	54.4 &	63.9 &	50.6 &	47.7 &	41.9 \\
+ CL (count-based) &  43.7	&	55.9	&	38.8	&	32.5	&	54.0	&	\bf 64.3	&	49.8	&	47.1	&	41.1 \\
+ CL (step-based) &  \bf 44.7 &  \bf 56.0 & \bf 40.1	&  \bf 33.3 &	 \bf 54.6 &	 64.2 &	\bf 50.8 &	 \bf 47.9 &	\bf 42.2 \\
        \midrule
MoE-64 \moetokendropoutabbrv{}	&	\bf 44.7	&	\bf 55.9	&	\bf 40.1	&	\bf 33.4	&	\bf 54.8	&	\bf 64.3	&	\bf 51.0	&	48.3	&	\bf 42.5	\\
+CL (step-based)	&	44.3	&	55.7	&	39.6	&	33.1	&	54.7	&	63.9	&	50.9	&	\bf 48.4	&	42.2	\\
    \bottomrule
    \end{tabular}}
    \caption{
    Results of Curriculum Learning applied to a vanilla MoE model and an MoE model with EOM. 
    }
    \label{tbl:modeling:cl}
\end{table}

\paragraph{Experimental Setup.}
To derive the phases of the curriculum, we train a vanilla MoE model with $\drop{=}0.3$ (our baseline), then, based on observed over-fitting patterns, we partition the tasks in our \ablation.
For both count and step-based curricula, we introduce pairs in $n{=}3$ phases over $\clupdates{=}100k$. 
For count-based curriculum, we partition language pairs into bins w.r.t. the training examples available for the task ($\mathcal D_t$): $b_1$ if $|\mathcal D_t| \geq 5e6$, 
$b_2$ if $8e5 \leq |\mathcal D_t| < 5e6$,
and $b_3$ if $|\mathcal D_t| < 8e5$. With that we use $(k_1,k_2,k_3)=(100k, 40k, 20k)$.\footnote{That means $b_1$ is introduced at step $0$, $b_2$ at step $\clupdates-40k$, and $b_3$ at step $\clupdates-20k$} 
For step-based curriculum, 
we follow \Cref{alg:cl:step} with $n=5$ and merge the first 3 buckets resulting in 3 bins introduced at $(k_1,k_2,k_3)=(100k, 40k, 20k)$. See \Cref{appendix:cl} for the exact partitioning.

To combine a stronger dropout regularization with Curriculum Learning methods, we next apply our best CL strategy (\textit{step-based}) to an MoE model with \moetokendropoutabbrv{} ($\moetokdrop{=}0.1$).

\paragraph{Results.}
We show the results of our CL experiments in \Cref{tbl:modeling:cl}. 
For the baseline MoE-64, by using \textit{step-based} CL, 
we improve the accuracy on very low-resource directions
by 0.8 \chrf{} in \langcode{eng-xx}
and 0.2 \chrf{} in \langcode{xx-eng} 
.
Across all resource levels, we improve the accuracy in \langcode{eng-xx} and \langcode{xx-eng} by 0.4 and 0.2 \chrf{}.
On non-English directions \textit{step-based} CL improves the quality by 0.3 \chrf{}.
The \textit{count-based} CL hurts the model performance in all tasks except from very low-resource \langcode{eng-xx} directions.

For MoE \moetokendropoutabbrv{}, training with \textit{step-based} CL actually hurts performance across all tasks except for \langcode{xx-eng} very low-resource. We hypothesize that over-fitting on our \ablation~is already reduced by \moetokendropoutabbrv{}, thus, adding a curriculum on top of that is not needed and has a negligible impact on translation quality. 

\section{Analysis of Multilingual Sparsely Gated MoE Models.}\label{sec:moe:analysis}
\begin{figure}[!t]
\centering
\includegraphics[width=.8\linewidth]{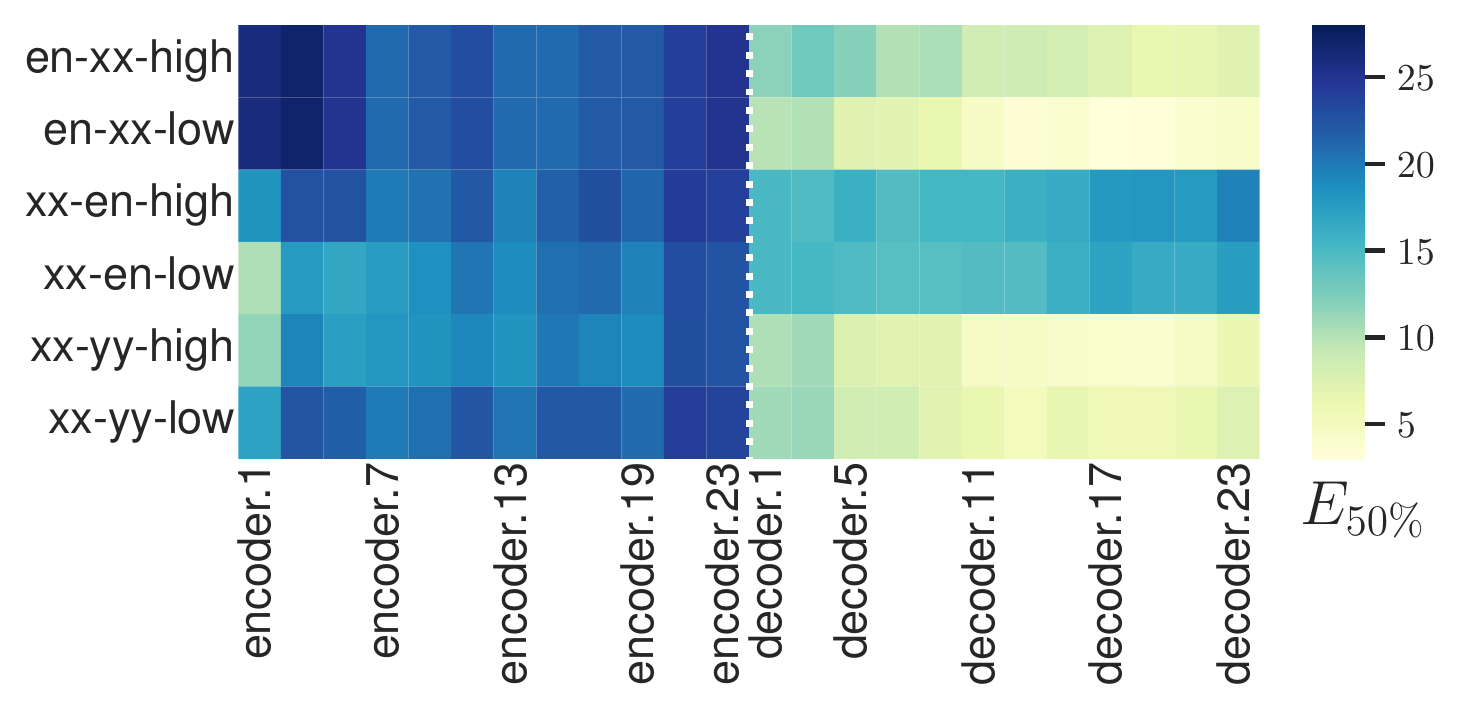}
\vspace{-5pt}
\caption{
$E_{50\%}$ for the $\drop{=}0.1$ model. 
Each column correspond to an MoE layer and we aggregate the statistic across 6 task groups.
}\label{fig:analysis:e50}
\end{figure}
\begin{figure*}[!ht]
\captionsetup[subfigure]{aboveskip=-1pt,belowskip=-1pt}
\centering
\begin{subfigure}[c]{.3\linewidth}
\includegraphics[height=2.0cm]{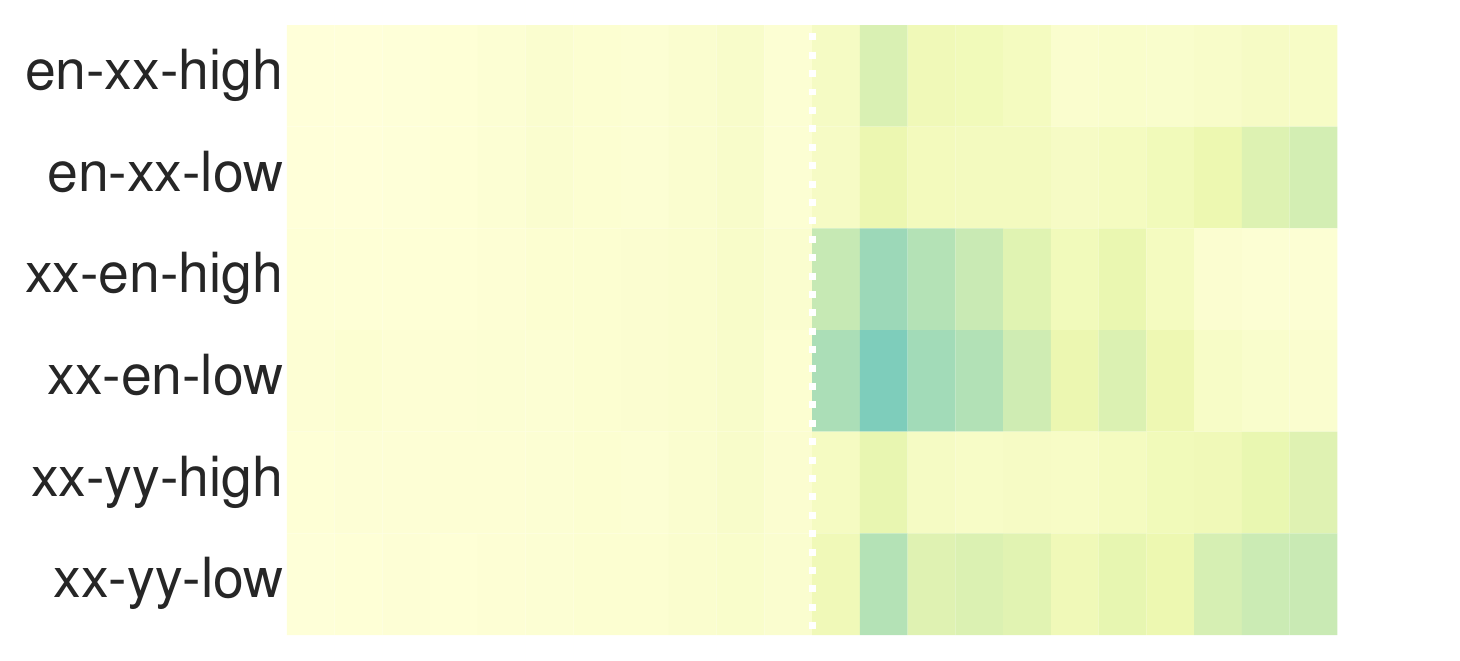}
\caption{$\drop{=}0.1$}
\end{subfigure}\hskip1pt
\begin{subfigure}[c]{.25\linewidth}
\includegraphics[height=2.0cm]{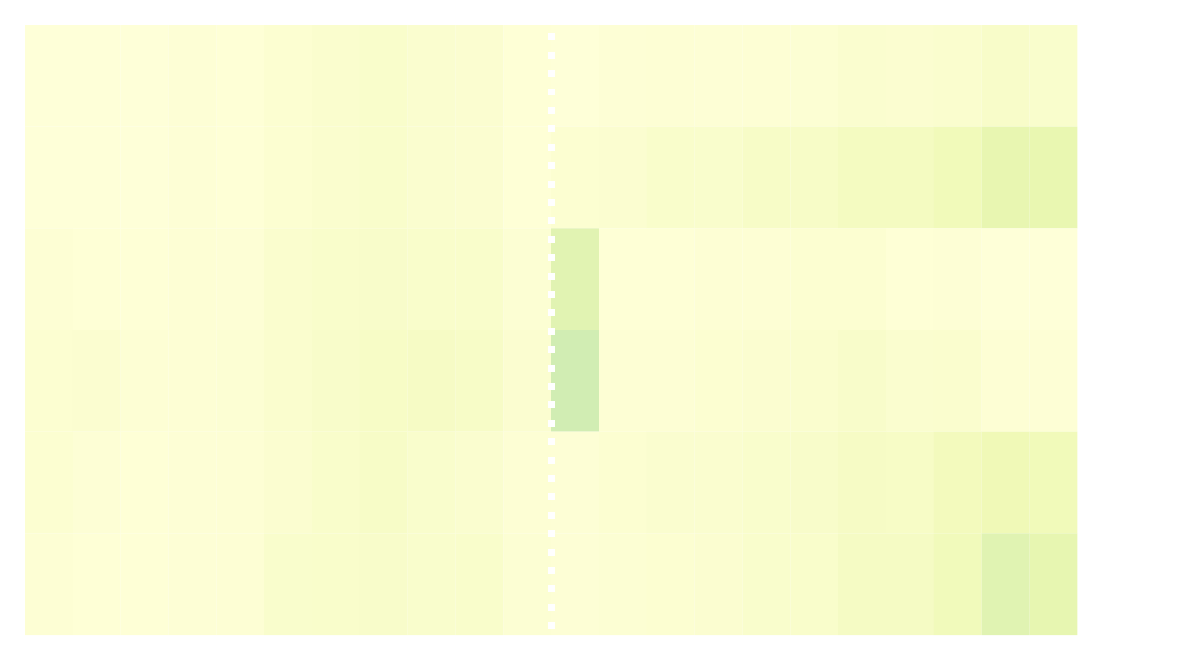}
\caption{$\drop{=}0.1$, $\moetokdrop{=}0.3$}
\end{subfigure}\hskip1pt
\begin{subfigure}[c]{.25\linewidth}
\includegraphics[height=2.0cm]{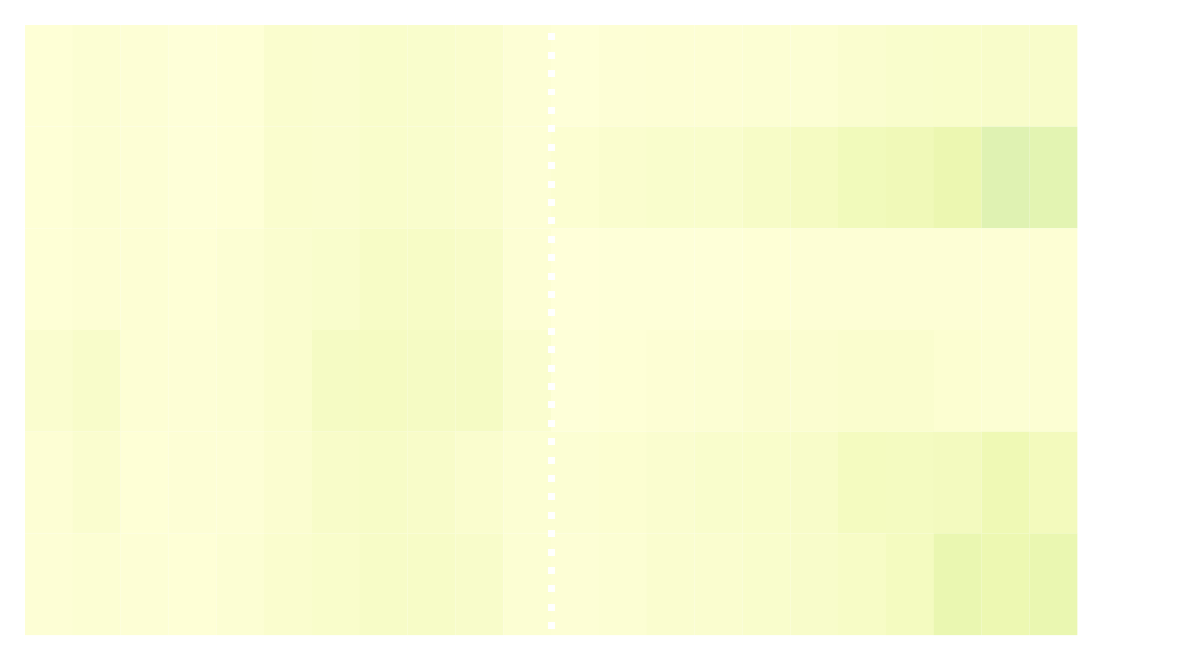}
\caption{$\drop{=}0.1$, $\fomdrop{=}0.3$}
\end{subfigure}\hfill\\
\begin{subfigure}[c]{.3\linewidth}
\includegraphics[height=2.9cm]{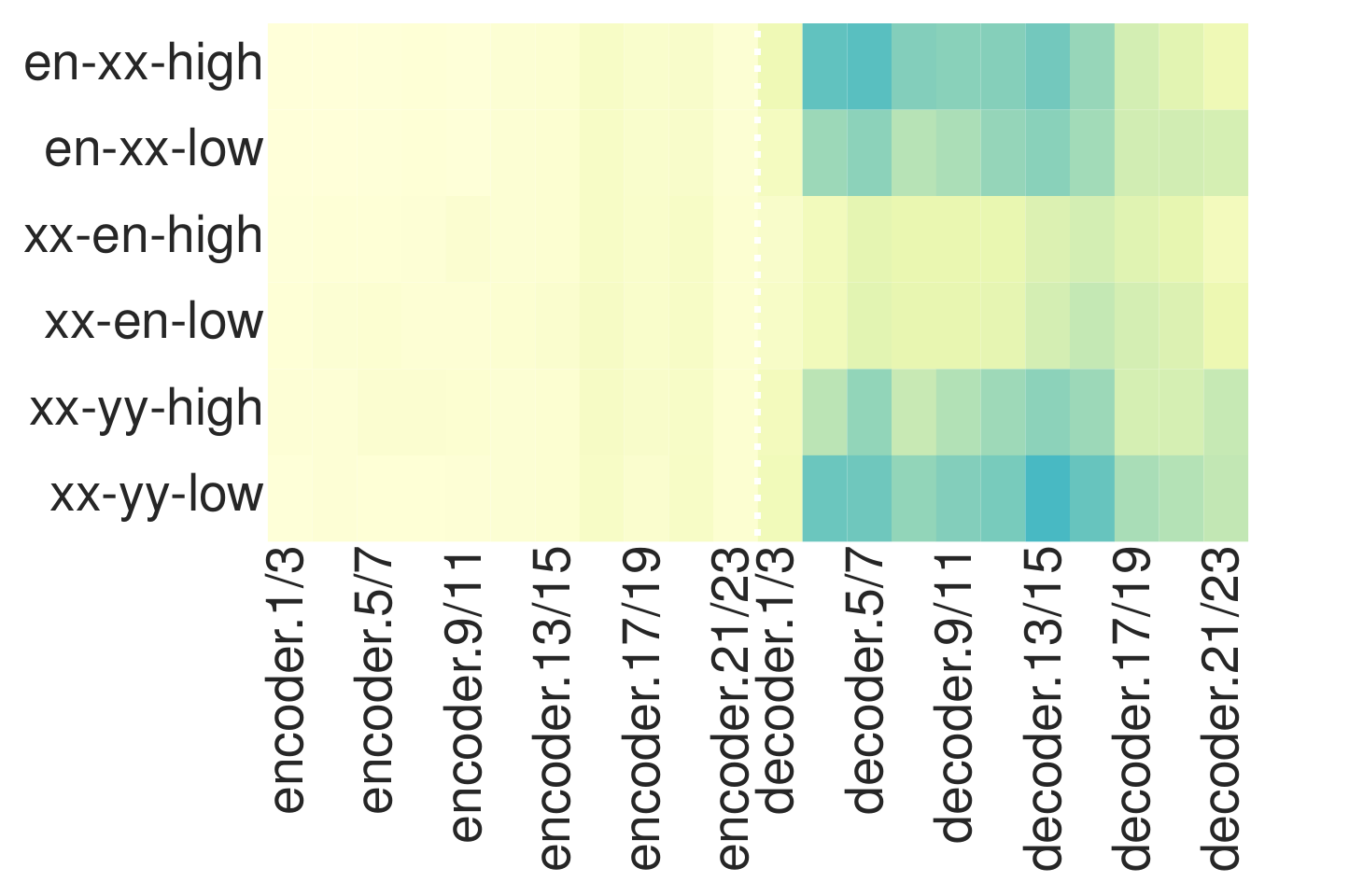}
\caption{$\drop{=}0.3$}
\end{subfigure}\hskip1pt
\begin{subfigure}[c]{.25\linewidth}
\includegraphics[height=2.9cm]{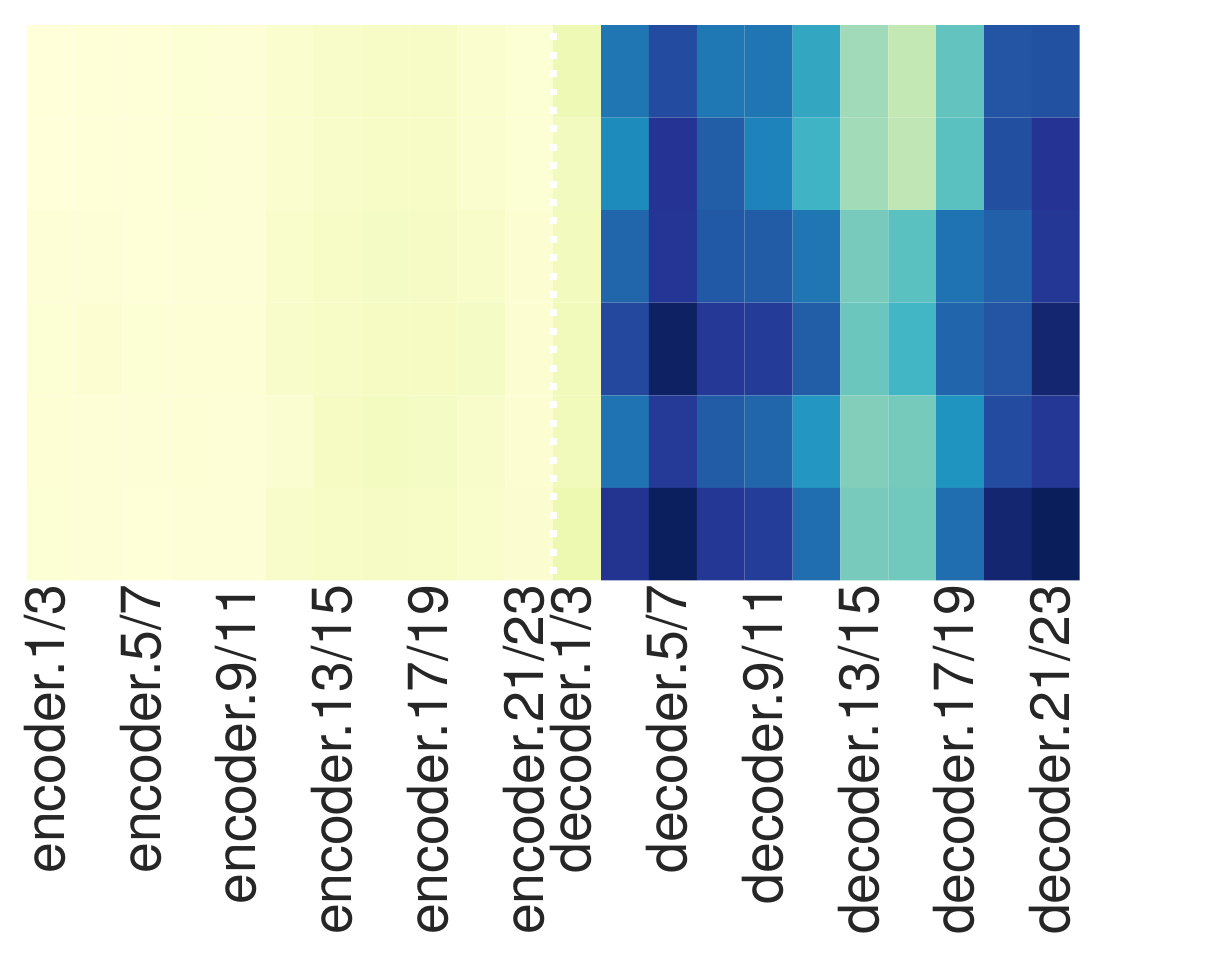}
\caption{$\drop{=}0.3$, $\moetokdrop{=}0.1$}
\end{subfigure}\hskip1pt
\begin{subfigure}[c]{.25\linewidth}
\includegraphics[height=2.9cm]{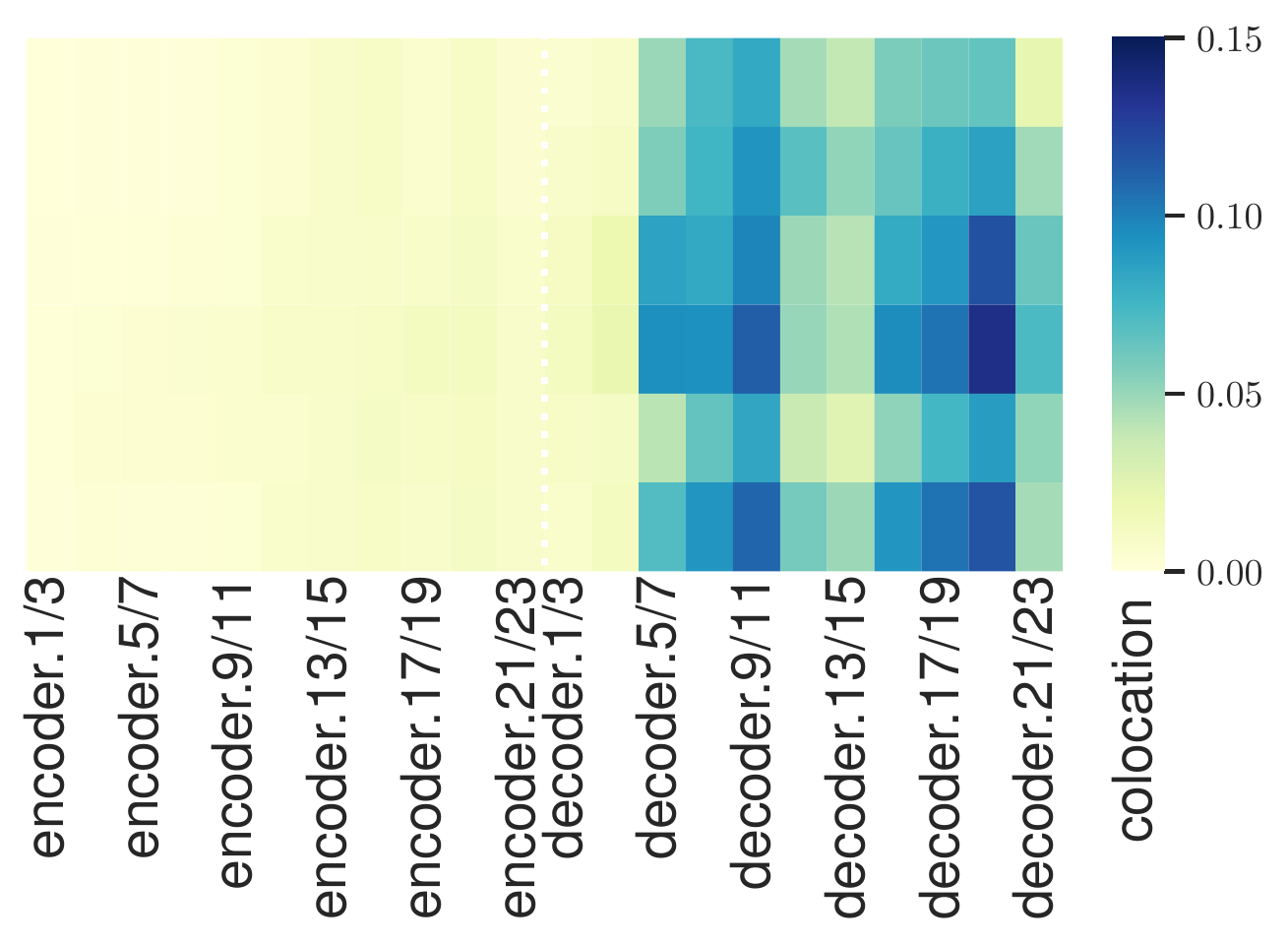}
\caption{$\drop{=}0.3$, $\fomdrop{=}0.1$}
\end{subfigure}\hfill\\
\vspace{-5pt}
\caption{Heatmap of experts' colocation in consecutive MoE layers of the encoder-deocder architecture. Darker colors correspond to higher colocation meaning that an input token routed to expert $i$ in layer $\ell$ will likely be routed to expert $j$ in layer ${\ell+1}$}
\label{fig:coloc:heatmap}
\end{figure*}

MoE theoretically enables models to specialize expert capacity for different tasks, but what do these models actually learn and how does our regularization methods affect these learnings?

To evaluate if experts are specialized in any given task (translation direction), we look at the experts' usage per task i.e., for a given MoE layer we report the distribution of the task tokens\footnote{source tokens for encoder layers and target tokens for decoder layers in \flores{}~\texttt{dev}} across the $E$ experts. For this metric, each token selects its top-1 expert.
The gating distributions are obtained in teacher-forcing mode, meaning that we feed the true target prefix to predict the next target token in the decoder side.
To summarize these distributions
we report the number of experts it take to cover 50\% of the tokens, denoted with $E_{50\%}$. A low $E_{50\%}$ means that a small number of experts is specialized in this task. We aggregate this statistic across 6 groups of tasks: English-to-many, Many-to-English and Non-English, with the suffix high or low for the resource level.

In \cref{fig:analysis:e50}, we show this statistic for the model trained with $\drop{=}0.1$\footnote{similar trends can be observed with other models, regardless of the regularization hyper-parameters.}
We observe that translating into English (xx-en-high/low) uses more experts in the decoder than translating into non-English (\langcode{xx-yy} and \langcode{en-xx}) and that low resource target languages (en-xx-low) activate less experts in the decoder.

To check if the specialized experts are only dedicated to a specific translation direction or if they are activated for any low-resource direction, we show in \Cref{fig:analysis:experts} of the appendix the heat-map of task experts for all 110 directions in our dataset for the $\drop{=}0.3$ model. We also look at the similarities in task experts and show in \Cref{fig:ablation:heatmap} of the appendix the similarities heat-maps at different MoE layers.

We next look in \Cref{fig:coloc:heatmap} at the co-location of experts in consecutive MoE layers, i.e., 
if tokens assigned to expert $i$ in $\text{layer}_\ell$ will automatically get routed to expert $j$ in $\text{layer}_{\ell+1}$
If $u_{ti}$ is token $x_t$'s usage of expert $i$, then the co-location  is evaluated as follows:
\begin{align}
\text{c}(\ell, {\ell+1})  = \max_{i,j}{\text{cor}_{x_t\sim \mathcal D}(u_{ti}^{\ell}, u_{tj}^{\ell+1})}
\end{align}
We observe that: (1) There is almost no co-location between the encoder’s MoE layers whereas the decoder’s MoE experts tend to co-locate. (2) $\drop{=}0.3$ leads to a higher co-location in the decoder’s MoE layers, paired with EOM, and to a lesser extent FOM, this co-location increases. (3) When the dropout rate is low (0.1), the addition of EOM or FOM weakens the collocation of the decoder’s experts.

We additionally use the drop in translation quality with random-routing in \Cref{app:random-routing} as a proxy for experts' specialization.
\section{Related work}
\paragraph{Improved routing in MoE models.}
Recent works have proposed alternatives to the commonly used top-2 gating of \citet{lepikhin2020gshard}:
Hash layers \citep{roller2021hash} use random fixed routing and
\citet{lewis2021base} view routing as a linear assignment problem and drop the load balancing loss. 
\citet{zuo2022taming} suggest to randomly select experts.
\citet{fedus2022switch} opt for top-1 routing, and
\citet{yang2021m6} split experts into different groups and applies k top-1 routing in each. 
In this work, we only use Top-2 gating\footnote{We did use top-1 gating for CMR to maintain a comparable computational cost with the baseline} but our techniques are orthogonal to the routing method.

\noindent
\paragraph{Regularizing MoE models.}
\citet{zoph2022designing} tried increasing the dropout within the expert (dubbed expert dropout) but saw marginal improvement in quality.
\citet{kim2021zcode} randomize the priority of tokens within a mini-batch as a regularization method.
\citet{liu2022gating} propose \textit{gating dropout} to reduce cross-machine communication in MoE layers.
\citet{xie2022moec} propose routing tokens to expert clusters and a cluster-level expert dropout. 

\noindent
\paragraph{Conditional compute and language-specific parameters.}
A common solution to relax parameter sharing in MMT models is to use light-weight language-specific adapters~\citep{rebuffi2017learning,bapna2019simple}. 
Their size, however, scales linearly in the number of languages.
\citet{baziotis2022multilingual} introduce hyper-adapters to generate the adapters themselves.
To make these language-specific parameters optional,~\citet{zhang2021clsr} propose CLSR to dynamically select language-specific or shared paths. These paths are simple linear projections and do not incorporate routing. Similar to our own CMR's budget loss, CLSR optimizes the MMT cross-entropy while constraining the use of the language-specific capacity.
Another approach similar to \clsrabbrv{}~is Residual-MoE~\citep{rajbhandari2022deepspeed}.
It is a hybrid dense and MoE model but it does not learn weights for each component.

\section{Conclusion}
In massively multilingual settings with imbalanced datasets, MoE models over-fit significantly more than dense models on low-resource directions. 
This work introduce multiple effective strategies for regularizing MoE models and achieving better performance across all language pairs, especially low-resource pairs. With \moetokendropoutabbrv{} and \fom, we propose dropout methods to further regularize MoE models. 
We introduce in \clsrabbrv~a novel architecture to balance the capacity between MoEs and shared dense paths.
Finally, we design curricula for introducing low-resource languages later during training. 
These strategies lead to less over-fitting on low-resource tasks, leading to improvements in translation quality.
\ifacl@finalcopy
\section*{Acknowledgments}
We would like to thank Philipp Koehn for his help with framing this paper. 
We would also like to thank James Cross, Onur Çelebi, Kevin Heffernan, Elahe Kalbassi, Janice Lam, Daniel Licht, Jean Maillard, Guillaume Wenzek, Bapi Akula, Loic Barrault, Gabriel Mejia Gonzalez, Prangthip Hansanti, Kaushik Ram Sadagopan, Pierre Andrews, Sergey Edunov, Angela Fan, Cynthia Gao, Vedanuj Goswami, Francisco Guzmán, Alexandre Mourachko, Christophe Ropers, Safiyyah Saleem, Holger Schwenk for their input on this work.
\fi

\bibliography{bibliography}

\begin{thebibliography}{32}
\expandafter\ifx\csname natexlab\endcsname\relax\def\natexlab#1{#1}\fi

\bibitem[{Almahairi et~al.(2016)Almahairi, Ballas, Cooijmans, Zheng,
  Larochelle, and Courville}]{almahairi:2016:dcn}
Amjad Almahairi, Nicolas Ballas, Tim Cooijmans, Yin Zheng, Hugo Larochelle, and
  Aaron Courville. 2016.
\newblock Dynamic capacity networks.
\newblock In \emph{Proceedings of the 33rd International Conference on
  International Conference on Machine Learning - Volume 48}, ICML'16, page
  2091–2100. JMLR.org.

\bibitem[{Ba et~al.(2016)Ba, Kiros, and Hinton}]{ba2016layer}
Lei~Jimmy Ba, Jamie~Ryan Kiros, and Geoffrey~E. Hinton. 2016.
\newblock \href {http://arxiv.org/abs/1607.06450} {Layer normalization}.
\newblock \emph{CoRR}, abs/1607.06450.

\bibitem[{Bapna and Firat(2019)}]{bapna2019simple}
Ankur Bapna and Orhan Firat. 2019.
\newblock Simple, scalable adaptation for neural machine translation.
\newblock In \emph{Proceedings of the 2019 Conference on Empirical Methods in
  Natural Language Processing and the 9th International Joint Conference on
  Natural Language Processing (EMNLP-IJCNLP)}, pages 1538--1548.

\bibitem[{Baziotis et~al.(2022)Baziotis, Artetxe, Cross, and
  Bhosale}]{baziotis2022multilingual}
Christos Baziotis, Mikel Artetxe, James Cross, and Shruti Bhosale. 2022.
\newblock Multilingual machine translation with hyper-adapters.
\newblock In \emph{EMNLP}.

\bibitem[{Bengio et~al.(2013)Bengio, L{\'{e}}onard, and
  Courville}]{bengio2013conditional}
Yoshua Bengio, Nicholas L{\'{e}}onard, and Aaron~C. Courville. 2013.
\newblock \href {http://arxiv.org/abs/1308.3432} {Estimating or propagating
  gradients through stochastic neurons for conditional computation}.
\newblock \emph{CoRR}, abs/1308.3432.

\bibitem[{Brown et~al.(2020)Brown, Mann, Ryder, Subbiah, Kaplan, Dhariwal,
  Neelakantan, Shyam, Sastry, Askell et~al.}]{brown2020language}
Tom Brown, Benjamin Mann, Nick Ryder, Melanie Subbiah, Jared~D Kaplan, Prafulla
  Dhariwal, Arvind Neelakantan, Pranav Shyam, Girish Sastry, Amanda Askell,
  et~al. 2020.
\newblock Language models are few-shot learners.
\newblock \emph{Advances in neural information processing systems},
  33:1877--1901.

\bibitem[{Du et~al.(2021)Du, Huang, Dai, Tong, Lepikhin, Xu, Krikun, Zhou, Yu,
  Firat, Zoph, Fedus, Bosma, Zhou, Wang, Wang, Webster, Pellat, Robinson,
  Meier{-}Hellstern, Duke, Dixon, Zhang, Le, Wu, Chen, and Cui}]{du2021glam}
Nan Du, Yanping Huang, Andrew~M. Dai, Simon Tong, Dmitry Lepikhin, Yuanzhong
  Xu, Maxim Krikun, Yanqi Zhou, Adams~Wei Yu, Orhan Firat, Barret Zoph, Liam
  Fedus, Maarten Bosma, Zongwei Zhou, Tao Wang, Yu~Emma Wang, Kellie Webster,
  Marie Pellat, Kevin Robinson, Kathy Meier{-}Hellstern, Toju Duke, Lucas
  Dixon, Kun Zhang, Quoc~V. Le, Yonghui Wu, Zhifeng Chen, and Claire Cui. 2021.
\newblock \href {http://arxiv.org/abs/2112.06905} {Glam: Efficient scaling of
  language models with mixture-of-experts}.
\newblock \emph{CoRR}, abs/2112.06905.

\bibitem[{Fedus et~al.(2022)Fedus, Zoph, and Shazeer}]{fedus2022switch}
William Fedus, Barret Zoph, and Noam Shazeer. 2022.
\newblock \href {http://jmlr.org/papers/v23/21-0998.html} {Switch transformers:
  Scaling to trillion parameter models with simple and efficient sparsity}.
\newblock \emph{Journal of Machine Learning Research}, 23(120):1--39.

\bibitem[{Heffernan et~al.(2022)Heffernan, {\c{C}}elebi, and
  Schwenk}]{heffernan2022bitext}
Kevin Heffernan, Onur {\c{C}}elebi, and Holger Schwenk. 2022.
\newblock Bitext mining using distilled sentence representations for
  low-resource languages.
\newblock In \emph{EMNLP}.

\bibitem[{Hwang et~al.(2022)Hwang, Cui, Xiong, Yang, Liu, Hu, Wang, Salas,
  Jose, Ram et~al.}]{hwang2022tutel}
Changho Hwang, Wei Cui, Yifan Xiong, Ziyue Yang, Ze~Liu, Han Hu, Zilong Wang,
  Rafael Salas, Jithin Jose, Prabhat Ram, et~al. 2022.
\newblock Tutel: Adaptive mixture-of-experts at scale.
\newblock \emph{arXiv preprint arXiv:2206.03382}.

\bibitem[{Kaplan et~al.(2020)Kaplan, McCandlish, Henighan, Brown, Chess, Child,
  Gray, Radford, Wu, and Amodei}]{kaplan2020scaling}
Jared Kaplan, Sam McCandlish, Tom Henighan, Tom~B. Brown, Benjamin Chess, Rewon
  Child, Scott Gray, Alec Radford, Jeffrey Wu, and Dario Amodei. 2020.
\newblock \href {http://arxiv.org/abs/2001.08361} {Scaling laws for neural
  language models}.
\newblock \emph{CoRR}, abs/2001.08361.

\bibitem[{Kim et~al.(2021)Kim, Awan, Muzio, Cruz{-}Salinas, Lu, Hendy,
  Rajbhandari, He, and Awadalla}]{kim2021zcode}
Young~Jin Kim, Ammar~Ahmad Awan, Alexandre Muzio, Andr{\'{e}}s~Felipe
  Cruz{-}Salinas, Liyang Lu, Amr Hendy, Samyam Rajbhandari, Yuxiong He, and
  Hany~Hassan Awadalla. 2021.
\newblock \href {http://arxiv.org/abs/2109.10465} {Scalable and efficient moe
  training for multitask multilingual models}.
\newblock \emph{CoRR}, abs/2109.10465.

\bibitem[{Kingma and Ba(2015)}]{kingma2015adam}
Diederik~P. Kingma and Jimmy Ba. 2015.
\newblock \href {http://arxiv.org/abs/1412.6980} {Adam: {A} method for
  stochastic optimization}.
\newblock In \emph{3rd International Conference on Learning Representations,
  {ICLR} 2015, San Diego, CA, USA, May 7-9, 2015, Conference Track
  Proceedings}.

\bibitem[{Kudo and Richardson(2018)}]{kudo2018sentencepiece}
Taku Kudo and John Richardson. 2018.
\newblock \href {https://doi.org/10.18653/v1/d18-2012} {Sentencepiece: {A}
  simple and language independent subword tokenizer and detokenizer for neural
  text processing}.
\newblock In \emph{Proceedings of the 2018 Conference on Empirical Methods in
  Natural Language Processing, {EMNLP} 2018: System Demonstrations, Brussels,
  Belgium, October 31 - November 4, 2018}, pages 66--71. Association for
  Computational Linguistics.

\bibitem[{Lepikhin et~al.(2020)Lepikhin, Lee, Xu, Chen, Firat, Huang, Krikun,
  Shazeer, and Chen}]{lepikhin2020gshard}
Dmitry Lepikhin, HyoukJoong Lee, Yuanzhong Xu, Dehao Chen, Orhan Firat, Yanping
  Huang, Maxim Krikun, Noam Shazeer, and Zhifeng Chen. 2020.
\newblock \href {http://arxiv.org/abs/2006.16668} {Gshard: Scaling giant models
  with conditional computation and automatic sharding}.
\newblock \emph{CoRR}, abs/2006.16668.

\bibitem[{Lewis et~al.(2021)Lewis, Bhosale, Dettmers, Goyal, and
  Zettlemoyer}]{lewis2021base}
Mike Lewis, Shruti Bhosale, Tim Dettmers, Naman Goyal, and Luke Zettlemoyer.
  2021.
\newblock Base layers: Simplifying training of large, sparse models.
\newblock In \emph{International Conference on Machine Learning}, pages
  6265--6274. PMLR.

\bibitem[{Liu et~al.(2022)Liu, Kim, Muzio, Mozafari, and
  Awadalla}]{liu2022gating}
Rui Liu, Young~Jin Kim, Alexandre Muzio, Barzan Mozafari, and Hany~Hassan
  Awadalla. 2022.
\newblock Gating dropout: Communication-efficient regularization for sparsely
  activated transformers.
\newblock \emph{arXiv preprint arXiv:2205.14336}.

\bibitem[{{NLLB Team} et~al.(2022){NLLB Team}, Costa-jussà, Cross, Çelebi,
  Elbayad, Heafield, Heffernan, Kalbassi, Lam, Licht, Maillard, Sun, Wang,
  Wenzek, Youngblood, Akula, Barrault, Mejia-Gonzalez, Hansanti, Hoffman,
  Jarrett, Sadagopan, Rowe, Spruit, Tran, Andrews, Ayan, Bhosale, Edunov, Fan,
  Gao, Goswami, Guzmán, Koehn, Mourachko, Ropers, Saleem, Schwenk, and
  Wang}]{nllb2022}
{NLLB Team}, Marta~R. Costa-jussà, James Cross, Onur Çelebi, Maha Elbayad,
  Kenneth Heafield, Kevin Heffernan, Elahe Kalbassi, Janice Lam, Daniel Licht,
  Jean Maillard, Anna Sun, Skyler Wang, Guillaume Wenzek, Al~Youngblood, Bapi
  Akula, Loic Barrault, Gabriel Mejia-Gonzalez, Prangthip Hansanti, John
  Hoffman, Semarley Jarrett, Kaushik~Ram Sadagopan, Dirk Rowe, Shannon Spruit,
  Chau Tran, Pierre Andrews, Necip~Fazil Ayan, Shruti Bhosale, Sergey Edunov,
  Angela Fan, Cynthia Gao, Vedanuj Goswami, Francisco Guzmán, Philipp Koehn,
  Alexandre Mourachko, Christophe Ropers, Safiyyah Saleem, Holger Schwenk, and
  Jeff Wang. 2022.
\newblock No language left behind: Scaling human-centered machine translation.
\newblock \emph{arXiv preprint arXiv:2207.04672}.

\bibitem[{Popovi\'{c}(2017)}]{popovic:2017:WMT}
Maja Popovi\'{c}. 2017.
\newblock \href {http://www.aclweb.org/anthology/W17-4770} {chrf++: words
  helping character n-grams}.
\newblock In \emph{Proceedings of the Second Conference on Machine Translation,
  Volume 2: Shared Task Papers}, pages 612--618, Copenhagen, Denmark.
  Association for Computational Linguistics.

\bibitem[{Rajbhandari et~al.(2022)Rajbhandari, Li, Yao, Zhang, Aminabadi, Awan,
  Rasley, and He}]{rajbhandari2022deepspeed}
Samyam Rajbhandari, Conglong Li, Zhewei Yao, Minjia Zhang, Reza~Yazdani
  Aminabadi, Ammar~Ahmad Awan, Jeff Rasley, and Yuxiong He. 2022.
\newblock Deepspeed-moe: Advancing mixture-of-experts inference and training to
  power next-generation ai scale.
\newblock \emph{arXiv preprint arXiv:2201.05596}.

\bibitem[{Rebuffi et~al.(2017)Rebuffi, Bilen, and
  Vedaldi}]{rebuffi2017learning}
Sylvestre-Alvise Rebuffi, Hakan Bilen, and Andrea Vedaldi. 2017.
\newblock Learning multiple visual domains with residual adapters.
\newblock \emph{Advances in neural information processing systems}, 30.

\bibitem[{Roller et~al.(2021)Roller, Sukhbaatar, Szlam, and
  Weston}]{roller2021hash}
Stephen Roller, Sainbayar Sukhbaatar, Arthur Szlam, and Jason~E Weston. 2021.
\newblock \href {https://openreview.net/forum?id=lMgDDWb1ULW} {Hash layers for
  large sparse models}.
\newblock In \emph{Advances in Neural Information Processing Systems}.

\bibitem[{Shazeer et~al.(2017)Shazeer, Mirhoseini, Maziarz, Davis, Le, Hinton,
  and Dean}]{shazeer2017outrageously}
Noam Shazeer, Azalia Mirhoseini, Krzysztof Maziarz, Andy Davis, Quoc Le,
  Geoffrey Hinton, and Jeff Dean. 2017.
\newblock \href {https://openreview.net/pdf?id=B1ckMDqlg} {Outrageously large
  neural networks: The sparsely-gated mixture-of-experts layer}.
\newblock In \emph{Proceedings of International Conference on Learning
  Representations (ICLR)}.

\bibitem[{Sutskever et~al.(2014)Sutskever, Vinyals, and Le}]{sutskever14nips}
Ilya Sutskever, Oriol Vinyals, and Quoc~V. Le. 2014.
\newblock \href
  {https://papers.nips.cc/paper/5346-sequence-to-sequence-learning-with-neural-networks}
  {Sequence to sequence learning with neural networks}.
\newblock In \emph{Proc. of NeurIPS}.

\bibitem[{Szegedy et~al.(2015)Szegedy, Vanhoucke, Ioffe, Shlens, and
  Wojna}]{szegedy:inception:2015}
Christian Szegedy, Vincent Vanhoucke, Sergey Ioffe, Jonathon Shlens, and
  Zbigniew Wojna. 2015.
\newblock \href {http://arxiv.org/abs/1512.00567} {Rethinking the inception
  architecture for computer vision}.
\newblock \emph{CoRR}, abs/1512.00567.

\bibitem[{Vaswani et~al.(2017)Vaswani, Shazeer, Parmar, Uszkoreit, Jones,
  Gomez, Kaiser, and Polosukhin}]{vaswani2017attention}
Ashish Vaswani, Noam Shazeer, Niki Parmar, Jakob Uszkoreit, Llion Jones,
  Aidan~N Gomez, {\L}ukasz Kaiser, and Illia Polosukhin. 2017.
\newblock Attention is all you need.
\newblock In \emph{Advances in neural information processing systems}, pages
  5998--6008.

\bibitem[{Xie et~al.(2022)Xie, Huang, Chen, and Wei}]{xie2022moec}
Yuan Xie, Shaohan Huang, Tianyu Chen, and Furu Wei. 2022.
\newblock Moec: Mixture of expert clusters.
\newblock \emph{arXiv preprint arXiv:2207.09094}.

\bibitem[{Xiong et~al.(2020)Xiong, Yang, He, Zheng, Zheng, Xing, Zhang, Lan,
  Wang, and Liu}]{xiong2020layer}
Ruibin Xiong, Yunchang Yang, Di~He, Kai Zheng, Shuxin Zheng, Chen Xing,
  Huishuai Zhang, Yanyan Lan, Liwei Wang, and Tieyan Liu. 2020.
\newblock On layer normalization in the transformer architecture.
\newblock In \emph{International Conference on Machine Learning}, pages
  10524--10533. PMLR.

\bibitem[{Yang et~al.(2021)Yang, Lin, Men, Zhou, Jiang, Jia, Wang, Zhang, Wang,
  Li et~al.}]{yang2021m6}
An~Yang, Junyang Lin, Rui Men, Chang Zhou, Le~Jiang, Xianyan Jia, Ang Wang, Jie
  Zhang, Jiamang Wang, Yong Li, et~al. 2021.
\newblock M6-t: Exploring sparse expert models and beyond.
\newblock \emph{arXiv preprint arXiv:2105.15082}.

\bibitem[{Zhang et~al.(2021)Zhang, Bapna, Sennrich, and Firat}]{zhang2021clsr}
Biao Zhang, Ankur Bapna, Rico Sennrich, and Orhan Firat. 2021.
\newblock \href {https://openreview.net/forum?id=Wj4ODo0uyCF} {Share or not?
  learning to schedule language-specific capacity for multilingual
  translation}.
\newblock In \emph{International Conference on Learning Representations}.

\bibitem[{Zoph et~al.(2022)Zoph, Bello, Kumar, Du, Huang, Dean, Shazeer, and
  Fedus}]{zoph2022designing}
Barret Zoph, Irwan Bello, Sameer Kumar, Nan Du, Yanping Huang, Jeff Dean, Noam
  Shazeer, and William Fedus. 2022.
\newblock St-moe: Designing stable and transferable sparse expert models.
\newblock \emph{arXiv preprint arXiv:2202.08906}.

\bibitem[{Zuo et~al.(2022)Zuo, Liu, Jiao, Kim, Hassan, Zhang, Gao, and
  Zhao}]{zuo2022taming}
Simiao Zuo, Xiaodong Liu, Jian Jiao, Young~Jin Kim, Hany Hassan, Ruofei Zhang,
  Jianfeng Gao, and Tuo Zhao. 2022.
\newblock \href {https://openreview.net/forum?id=B72HXs80q4} {Taming sparsely
  activated transformer with stochastic experts}.
\newblock In \emph{International Conference on Learning Representations}.

\end{thebibliography}
\bibliographystyle{acl_natbib}

\clearpage
\appendix
\section{Training data}\label{appendix:data}
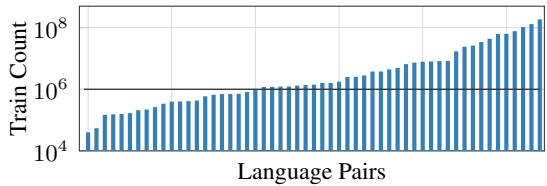
\begin{figure}[!h]
\centering
\begin{tikzpicture}
    \begin{axis}[
            grid=both,
            xlabel={Language Pairs},
            ylabel={Train Count},
            ymin = 10000,
            ymode=log,
            xticklabels={,,},
            xtick style={draw=none},
            height=3.5cm,
            width=\linewidth,
            enlarge x limits=0.01,
            x label style={font=\small, yshift=7pt},
            y label style={font=\small, yshift=-5pt},
            tick label style={font=\small,},
            ]
        
        \addplot[ybar,bar width=0.5, draw=none, fill=Set1B] coordinates {
(0, 39992)
(1, 54534)
(2, 147886)
(3, 153710)
(4, 158554)
(5, 168631)
(6, 208656)
(7, 219738)
(8, 263496)
(9, 339901)
(10, 397535)
(11, 402450)
(12, 412513)
(13, 432823)
(14, 593767)
(15, 664013)
(16, 697681)
(17, 700642)
(18, 711352)
(19, 811403)
(20, 1009697)
(21, 1185770)
(22, 1198193)
(23, 1223244)
(24, 1234763)
(25, 1314657)
(26, 1367592)
(27, 1409641)
(28, 1607268)
(29, 1608456)
(30, 1763488)
(31, 2508503)
(32, 2537255)
(33, 2808689)
(34, 3771110)
(35, 3786315)
(36, 4434968)
(37, 4952193)
(38, 6536525)
(39, 7329369)
(40, 7859748)
(41, 8031606)
(42, 8199306)
(43, 8338338)
(44, 17024673)
(45, 24118783)
(46, 26391896)
(47, 34071448)
(48, 43694544)
(49, 62797105)
(50, 63119065)
(51, 76920466)
(52, 104650121)
(53, 131314156)
(54, 186633573)
    };
\addplot[smooth, darkgray, line width=0.5pt] coordinates {(-0.5,1000000) (54, 1000000)};
\end{axis}
\end{tikzpicture}
\vspace{-5pt}
\caption{Training data across all language pairs in our \ablation.
}\label{fig:modeling:ablation_dataset_counts}
\end{figure}

We list in \Cref{tab:datacounts} the amount of data (bitexts) used to train our models. 
\Cref{fig:modeling:ablation_dataset_counts} shows the data distribution over language pairs sorted by the example count per pair. The highest resource language pair has 180M examples (English-French), and the lowest resource language pair has 40K examples (Hindi-Tamil). 

\section{Training details}\label{appendix:training_hparams}
We train Transformer encoder-decoder models with dimension \modHiddenDim{}, FFN dimension \modFFNDim{}, \modAttHeads{} attention heads, \modLayers{} encoder layers and \modLayers{} decoder layers. 
We apply Layer-normalization~\citep{ba2016layer} at the beginning of each Transformer sub-layer (Pre-LN), as opposed to after the residual connection (Post-LN). 
This is because Pre-LN is more stable in practice compared to Post-LN~\citep{xiong2020layer}.
All models are trained for 100k updates with an effective batch size of 1M tokens per update.
We optimize with Adam~\citep{kingma2015adam} using ($\beta_1, \beta_2, \epsilon) = (0.9, 0.98, 10^{-6})$.
We linearly increase the learning rate up to 0.004 through 8000 warmup updates, then follow the inverse square root learning rate schedule.
For \topgating{2}, we set the expert capacity to $2\times T/E$, i.e., we enforce that each expert processes, at most, $2\times T/E$ tokens, where $T$ is the number of tokens in the mini-batch and $E$ is the number of experts.
During generation, we set the capacity to $T$ so that all tokens can be routed to whichever expert they choose.

\begin{table*}
\small
\centering
\begin{tabular}[t]{ll}
\toprule
code & language \\
\midrule
ace\_Latn & Acehnese\\
afr & Afrikaans\\
ara\_Arab & Arabic\\
ast & Asturian\\
ayr & Aymara\\
bel & Belarussian \\
bul & Bulgarian\\
cjk & Chokwe\\
cym & Welsh\\
eus & Basque\\
ewe & Ewe\\
fas & Persian\\
fin & Finnish\\
fon & Fon\\
fra & French\\
fuv & Fula\\
hau & Hausa\\
hin & Hindi\\
isl & Icelandic\\
ita & Italian\\
jpn & Japanese\\
kea & Kabuverdianu\\
kik & Kikuyu\\
kin & Kinyarwanda\\
kon & Kongo \\
kor & Korean\\
lav & Latvian\\
lin & Lingala\\
luo & Luo\\
mal & Malayalam\\
mar & Marathi\\
nso & Northern Sotho\\
oci & Occitan\\
por & Portuguese\\
run & Rundi\\
rus & Russian\\
sin & Sinhalese\\
snd & Sindhi\\
swh & Swahili\\
tam & Tamil\\
tat\_Cyrl & Tatar\\
tel & Telugu\\
tir & Tigrinya\\
tsn & Tswana\\
tso & Tsonga\\
twi & Twi\\
urd & Urdu\\
vie & Vietnamese\\
wol & Wolof\\
yor & Yoruba\\
yue & Yue Chinese\\
zho\_Hans & Chinese\\
\bottomrule
\end{tabular}\hskip8pt
\begin{tabular}[t]{lHrr}
\toprule
direction &   count & primary & mined \\
\midrule
eng-ace\_Latn & 1,185,350 & 36,591 & 1,148,759 \\
eng-afr & 7,289,928 & 1,449,916 & 5,840,012 \\
eng-ara\_Arab & 75,788,802 & 36,340,863 & 39,447,939 \\
eng-ast & 875,410 & 526 & 874,884 \\
eng-ayr & 679,934 & 69,185 & 610,749 \\
eng-bel & 939,643 & 47,166 & 892,477 \\
eng-bul & 62,448,652 & 26,706,641 & 35,742,011 \\
eng-cjk & 693,442 & 33,038 & 660,404 \\
eng-cym & 4,389,062 & 149,598 & 4,239,464 \\
eng-ewe & 1,273,925 & 534,793 & 739,132 \\
eng-fas & 23,930,039 & 4,402,104 & 19,527,935 \\
eng-fin & 62,027,853 & 34,784,117 & 27,243,736 \\
eng-fon & 335,817 & 36,752 & 299,065 \\
eng-fra & 179,922,947 & 37,993,938 & 141,929,009 \\
eng-fuv & 207,917 & 18,242 & 189,675 \\
eng-hau & 4,944,179 & 345,481 & 4,598,698 \\
eng-hin & 26,186,500 & 1,688,720 & 24,497,780 \\
eng-isl & 7,840,462 & 1,096,312 & 6,744,150 \\
eng-ita & 127,437,187 & 44,712,431 & 82,724,756 \\
eng-kea & 150,981 & 4,727 & 146,254 \\
eng-kik & 218,136 & 98,740 & 119,396 \\
eng-kin & 2,800,387 & 376,914 & 2,423,473 \\
eng-kon & 402,050 & 188,251 & 213,799 \\
eng-lav & 14,566,938 & 3,867,869 & 10,699,069 \\
eng-lin & 1,221,481 & 666,273 & 555,208 \\
eng-luo & 799,367 & 129,000 & 670,367 \\
eng-mal & 8,288,573 & 585,452 & 7,703,121 \\
eng-mar & 6,478,501 & 335,259 & 6,143,242 \\
eng-nso & 1,170,683 & 526,097 & 644,586 \\
eng-oci & 591,732 & 5,915 & 585,817 \\
eng-run & 1,593,139 & 454,678 & 1,138,461 \\
eng-rus & 101,477,342 & 30,271,773 & 71,205,569 \\
eng-sin & 3,750,000 & 461,857 & 3,288,143 \\
eng-snd & 2,529,730 & 95,718 & 2,434,012 \\
eng-tam & 7,904,241 & 680,297 & 7,223,944 \\
eng-tel & 8,134,423 & 253,718 & 7,880,705 \\
eng-tir & 1,212,898 & 83,980 & 1,128,918 \\
eng-tso & 1,592,993 & 711,883 & 881,110 \\
eng-twi & 1,729,025 & 508,746 & 1,220,279 \\
eng-urd & 3,748,179 & 875,172 & 2,873,007 \\
eng-vie & 43,472,533 & 3,689,843 & 39,782,690 \\
eng-wol & 156,979 & 9,233 & 147,746 \\
eng-yor & 2,496,961 & 397,793 & 2,099,168 \\
eng-yue & 54,534 & 54,534 & 0 \\
eng-zho\_Hans & 33,913,340 & 228,658 & 33,684,682 \\
\bottomrule
\end{tabular}\hskip8pt
\begin{tabular}[t]{lHrr}
\toprule
direction &     count & primary & mined\\
\midrule
ara\_Arab-sin & 402,450 & 402,450 & 0 \\
eus-por & 432,823 & 432,823 & 0 \\
fra-hau & 168,631 & 168,631 & 0 \\
fra-kon & 147,886 & 147,886 & 0 \\
fra-lin & 397,535 & 397,535 & 0 \\
fra-swh & 664,013 & 664,013 & 0 \\
hin-tam & 39,992 & 39,992 & 0 \\
jpn-kor & 1,009,697 & 1,009,697 & 0 \\
rus-tat\_Cyrl & 263,496 & 263,496 & 0 \\
swh-tsn & 697,681 & 697,681 & 0 \\
\bottomrule
\end{tabular}
\caption{List of languages and Data counts between primary (pre-existing publicly available parallel data) and mined~\citep{heffernan2022bitext} for the 110 directions of our \ablation. 45 languages are paired with English for a total of 90 English-centric directions. The remaining 20 directions are non-English centric.}
\label{tab:datacounts}
\end{table*}

\section{Curriculum Learning}\label{appendix:cl}

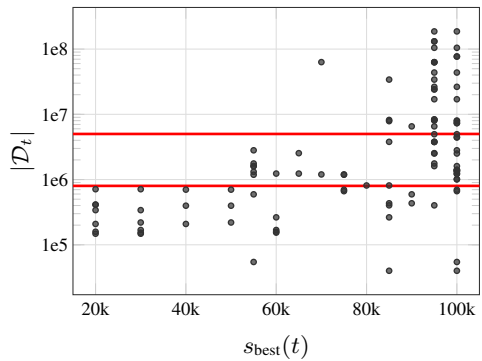
\begin{figure}[!ht]
\centering
\begin{tikzpicture}[scale=1]
\def\xmin{15000}
\def\xmax{105000}

\begin{groupplot}[
        group style={group size=2 by 1, horizontal sep=15pt, vertical sep=20pt},
        grid=major,
        label style={font=\small},
        y label style={yshift=-5pt},
        y tick label style={font=\scriptsize, xshift=2pt},
        x tick label style={font=\scriptsize, yshift=1pt},
        ymode=log,
        scaled x ticks=false,
        y axis line style=-,
        height=0.7\columnwidth,
        width=0.9\columnwidth,
        xtick={20000,40000,...,100000},
        xticklabels={20k,40k,60k,80k,100k},
        xmin=\xmin,xmax=\xmax,
        ytick={1e5, 1e6, 1e7, 1e8},
        yticklabels={1e5, 1e6, 1e7, 1e8},
       ]
\nextgroupplot[
    xlabel={$s_\text{best}(t)$},%
    ylabel={$|\mathcal D_t|$},%
    legend style={at={(0.01,0.01)},anchor=south west, font=\small, nodes={scale=0.8, transform shape}}
   ]

   \addplot+[darkgray, only marks, mark size=1pt, mark=*, line width=0.5pt, fill opacity=0.7]
    table [y=examples,x=step]{results/ablation/naive_cl_steps.dat};
    \addplot[mark=none, red, line width=1pt] coordinates {(\xmin, 8e5) (\xmax, 8e5)};
    \addplot[mark=none, red, line width=1pt] coordinates {(\xmin, 5e6) (\xmax, 5e6)};

\end{groupplot}
\end{tikzpicture}
\caption{For the baseline MoE model, we plot the steps corresponding to the best validation perplexity ($s_\text{best}$ on the x-axis) against the number of training examples ($|\mathcal D_t|$ on the y-axis).}
\label{fig:modeling:cl_naive_plot}
\end{figure}

\paragraph{\textit{Count-based} CL.} 
We empirically partition based on training example counts.
We first train our baseline model (MoE-64 $(\drop{=}0.3$) without CL, then we look at possible correlations between the number of steps before over-fitting and the count of training examples. In \Cref{fig:modeling:cl_naive_plot} we plot these data points with the counts on the y-axis and the start-of-over-fitting step on the x-axis. The horizontal red lines indicate where the \textit{count-based} curriculum thresholds were set in order to partition language pairs into bins.

We list in \Cref{tab:appendix:cl:count:baseline} the tasks in each bin for the baseline MoE model.

\paragraph{\textit{Step-based} CL.} 
We partition based on the step where we observed a task to start over-fitting. Following \Cref{alg:cl:step}, we partition the tasks into $n$ bins.
In our experiments, we started with $n{=}5$ resulting in a $\Delta$ of 20k steps. However, we merged the first three bins with characteristic steps $k_1=100k$, $k_2=80k$ and $k_3=60k$ to remain comparable with \textit{count-based} CL.

\begin{table}[!ht]
\small
\resizebox{\columnwidth}{!}{%
\begin{tabular}{lllp{5cm}}
    \toprule
     bin $b_i$ & \#tasks & $k_i$ & Language pairs  \\
     \midrule
     $b_1$ & 17$\times$2 & 100k &
     eng-afr,
     eng-ara\_Arab,
     eng-bul,
     eng-fas,
     eng-fin,
     eng-fra,
     eng-hin,
     eng-isl,
     eng-ita,
     eng-lav,
     eng-mal,
     eng-mar,
     eng-rus,
     eng-tam,
     eng-tel,
     eng-vie,
     eng-zho\_Hans
     \\
     \midrule
    $b_2$ & 24$\times$2 & 40k &
    eng-ace\_Latn,
    eng-ast,
    eng-ayr,
    eng-bel,
    eng-cjk,
    eng-cym,
    eng-ewe,
    eng-hau,
    eng-kin,
    eng-lin,
    eng-luo,
    eng-nso,
    eng-oci,
    eng-run,
    eng-sin,
    eng-snd,
    eng-tir,
    eng-tso,
    eng-twi,
    eng-urd,
    eng-yor,
    fra-swh,
    jpn-kor,
    swh-tsn \\
     \midrule
     $b_3$ & 14$\times$2 & 20k &
    eng-fon,
    eng-fuv,
    eng-kea,
    eng-kik,
    eng-kon,
    eng-wol,
    eng-yue,
    ara\_Arab-sin,
    eus-por,
    fra-hau,
    fra-kon,
    fra-lin,
    hin-tam,
    rus-tat\_Cyrl
     \\
     \bottomrule
\end{tabular}
}
\caption{\textit{Count-based} CL bins for the baseline MoE model ($\drop{=}0.3$). Step represents the number of steps the language pairs in this bin are trained}
\label{tab:appendix:cl:count:baseline}
\end{table}
\begin{table}[!ht]
\small
\resizebox{\columnwidth}{!}{%
\begin{tabular}{lllp{5cm}}
    \toprule
    bin $b_i$ & \#tasks & $k_i$ & Language pairs  \\
    \midrule
    $b_1$ & 86 & 100k &
    ace\_Latn-eng,
    afr-eng,
    ara\_Arab-eng,
    ara\_Arab-sin,
    ast-eng,
    bel-eng,
    bul-eng,
    cym-eng,
    eng-afr,
    eng-ara\_Arab,
    eng-ast,
    eng-bel,
    eng-bul,
    eng-cym,
    eng-ewe,
    eng-fas,
    eng-fin,
    eng-fra,
    eng-hau,
    eng-hin,
    eng-isl,
    eng-ita,
    eng-kea,
    eng-kin,
    eng-lav,
    eng-luo,
    eng-mal,
    eng-mar,
    eng-nso,
    eng-oci,
    eng-run,
    eng-rus,
    eng-sin,
    eng-snd,
    eng-tam,
    eng-tel,
    eng-tir,
    eng-tso,
    eng-twi,
    eng-urd,
    eng-vie,
    eng-yor,
    eng-zho\_Hans,
    eus-por,
    ewe-eng,
    fas-eng,
    fin-eng,
    fra-eng,
    fra-hau,
    fra-swh,
    hau-eng,
    hin-eng,
    hin-tam,
    isl-eng,
    ita-eng,
    jpn-kor,
    kin-eng,
    kor-jpn,
    lav-eng,
    lin-eng,
    lin-fra,
    luo-eng,
    mal-eng,
    mar-eng,
    nso-eng,
    por-eus,
    rus-eng,
    sin-ara\_Arab,
    sin-eng,
    snd-eng,
    swh-fra,
    swh-tsn,
    tam-eng,
    tam-hin,
    tat\_Cyrl-rus,
    tel-eng,
    tir-eng,
    tsn-swh,
    tso-eng,
    twi-eng,
    urd-eng,
    vie-eng,
    wol-eng,
    yor-eng,
    yue-eng,
    zho\_Hans-eng \\
     \midrule
    $b_2$ & 12 & 40k &
    ayr-eng,
    cjk-eng,
    eng-ace\_Latn,
    eng-ayr,
    eng-kik,
    eng-lin,
    fra-lin,
    fuv-eng,
    kik-eng,
    oci-eng,
    run-eng,
    rus-tat\_Cyrl \\
     \midrule
     $b_3$ & 12 & 20k &
    eng-cjk,
    eng-fon,
    eng-fuv,
    eng-kon,
    eng-wol,
    eng-yue,
    fon-eng,
    fra-kon,
    hau-fra,
    kea-eng,
    kon-eng,
    kon-fra
     \\
     \bottomrule
\end{tabular}
}
\caption {\textit{Step-based} CL bins for the baseline MoE-64 ($\moetokdrop{=}0.1$)}
\label{tab:appendix:cl:step:baseline}
\end{table}
\begin{table}[!ht]
\small
\resizebox{\columnwidth}{!}{%
\begin{tabular}{lllp{5cm}}
    \toprule
    bin $b_i$ & \#tasks & $k_i$ & Language pairs  \\
    \midrule
    $b_1$ & 95 &  100k &
    ace\_Latn-eng,
    afr-eng,
    ara\_Arab-eng,
    ara\_Arab-sin,
    ast-eng,
    ayr-eng,
    bel-eng,
    bul-eng,
    cym-eng,
    eng-ace\_Latn,
    eng-afr,
    eng-ara\_Arab,
    eng-ast,
    eng-bel,
    eng-bul,
    eng-cym,
    eng-ewe,
    eng-fas,
    eng-fin,
    eng-fra,
    eng-hau,
    eng-hin,
    eng-isl,
    eng-ita,
    eng-kea,
    eng-kik,
    eng-kin,
    eng-lav,
    eng-luo,
    eng-mal,
    eng-mar,
    eng-nso,
    eng-oci,
    eng-run,
    eng-rus,
    eng-sin,
    eng-snd,
    eng-tam,
    eng-tel,
    eng-tir,
    eng-tso,
    eng-twi,
    eng-urd,
    eng-vie,
    eng-yor,
    eng-zho\_Hans,
    eus-por,
    ewe-eng,
    fas-eng,
    fin-eng,
    fra-eng,
    fra-hau,
    fra-lin,
    fra-swh,
    fuv-eng,
    hau-eng,
    hau-fra,
    hin-eng,
    hin-tam,
    isl-eng,
    ita-eng,
    jpn-kor,
    kin-eng,
    kor-jpn,
    lav-eng,
    lin-eng,
    lin-fra,
    luo-eng,
    mal-eng,
    mar-eng,
    nso-eng,
    oci-eng,
    por-eus,
    run-eng,
    rus-eng,
    rus-tat\_Cyrl,
    sin-ara\_Arab,
    sin-eng,
    snd-eng,
    swh-fra,
    swh-tsn,
    tam-eng,
    tam-hin,
    tat\_Cyrl-rus,
    tel-eng,
    tir-eng,
    tsn-swh,
    tso-eng,
    twi-eng,
    urd-eng,
    vie-eng,
    wol-eng,
    yor-eng,
    yue-eng,
    zho\_Hans-eng \\
     \midrule
    $b_2$ & 5 & 40k &
    eng-ayr,
    eng-cjk,
    eng-lin,
    eng-wol,
    eng-yue \\
     \midrule
     $b_3$ & 10 & 20k &
    cjk-eng,
    eng-fon,
    eng-fuv,
    eng-kon,
    fon-eng,
    fra-kon,
    kea-eng,
    kik-eng,
    kon-eng,
    kon-fra
     \\
     \bottomrule
\end{tabular}
}
\caption {\textit{Step-based} CL bins for the baseline MoE-64 \moetokendropoutabbrv{} 
 ($\drop{=}0.3$,, $\moetokdrop{=}0.1$)}
\label{tab:appendix:cl:step:eom}
\end{table}

\section{Analysis of Multilingual Sparsely Gated MoE Models}
\subsection{Task Experts}
\begin{figure*}[!ht]
\centering
\includegraphics[width=\textwidth]{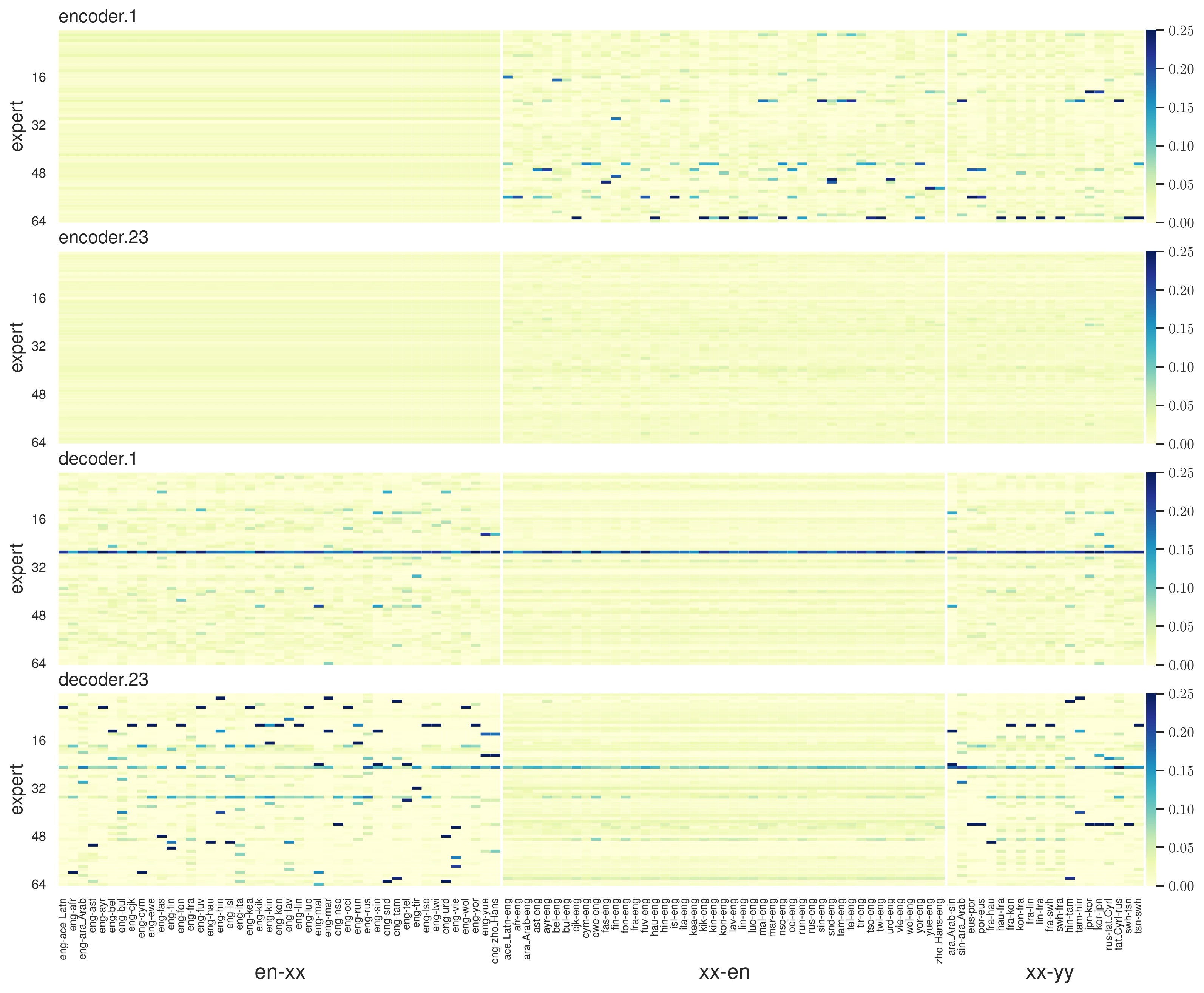}
\caption{Task experts for all 110 direction of our \ablation. Each subplot corresponds to a specific MoE layer and each column of the subplot show the experts' usage in the specific MoE layer for a translation direction. We show the translation directions in 3 groups: English-to-many (en-xx), many-to-English (xx-en) and Non-English (xx-yy).}
\label{fig:analysis:experts}
\end{figure*}

We show in \cref{fig:analysis:experts} the task experts for all 110 directions in our \ablation. These statistics correspond to the MoE-64 $\drop{=}0.3$ model but similar trends are observed for all other 5 models we analyzed.
Predictably, we observe that in English-to-many directions (en-xx, left side of the figure) the encoder experts are almost uniformly activated whereas the encoder experts are specialized per target language. The opposite can observed to a lesser extent on many-to-English direction (xx-en); we see some variation in decoder layers, but most importantly we see little variation in the late encoder layers (encoder.23) likely because the source representation at that stage is becoming more-or-less language-agnostic.

\subsection{Language experts.}

To obtain statistics on a language's usage of experts, we combine tokens from tasks that include our language of interest, $\lang$, before computing the density histogram of the top-1 experts. 
When looking at encoder MoE layers, we consider $\lang{\to}x$ tasks and when looking at decoder MoE layers we consider $x{\to}\lang$ tasks.

\begin{figure*}[!ht]
\centering
\begin{subfigure}[c]{.45\textwidth}
\includegraphics[width=\textwidth]{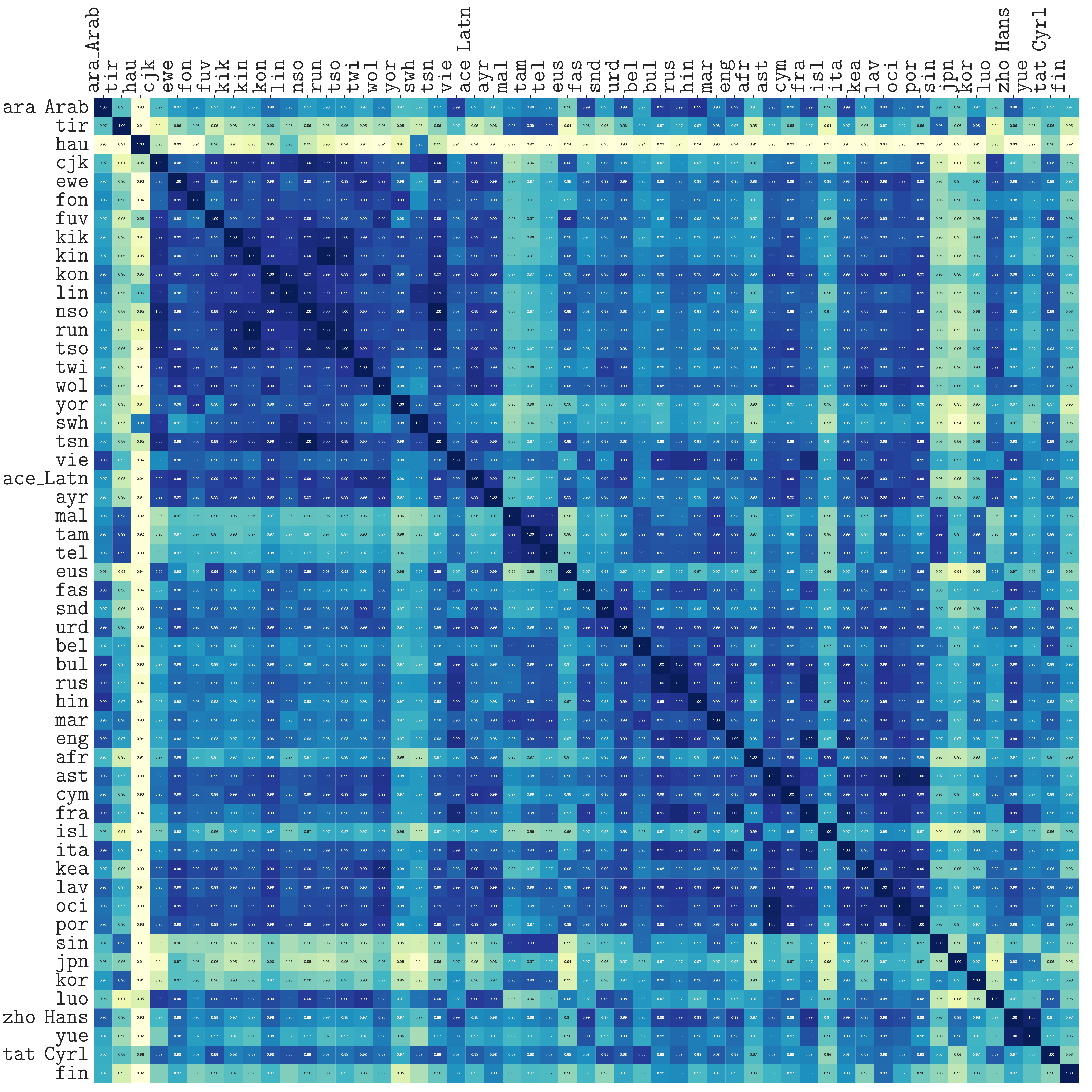}
\caption{First encoder MoE layer}\label{fig:ablation:heatmap:enc1}
\end{subfigure}
\begin{subfigure}[c]{.45\textwidth}
\includegraphics[width=\textwidth]{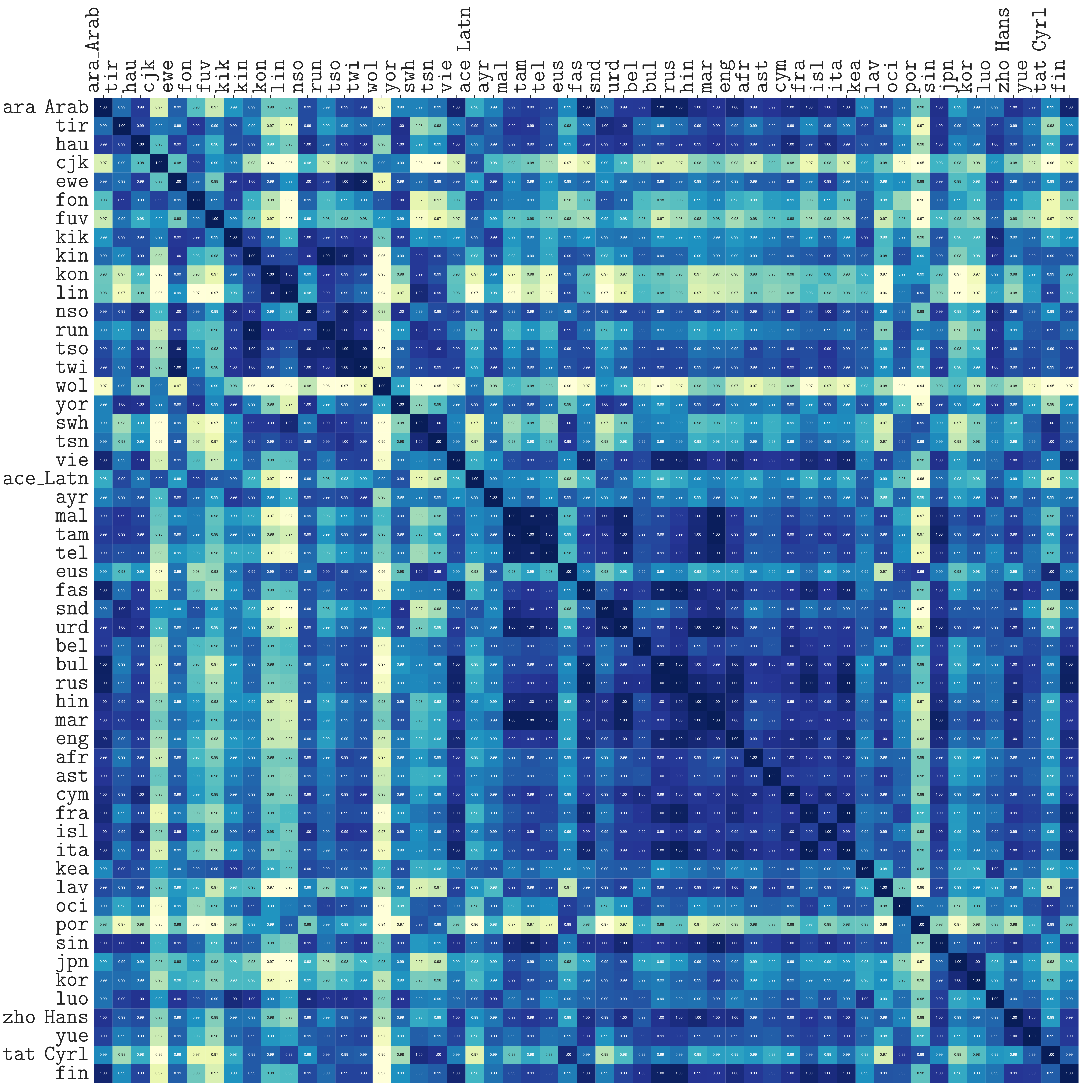}
\caption{Last encoder MoE layer}\label{fig:ablation:heatmap:enc23}
\end{subfigure}\\
\begin{subfigure}[c]{.45\textwidth}
\includegraphics[width=\textwidth]{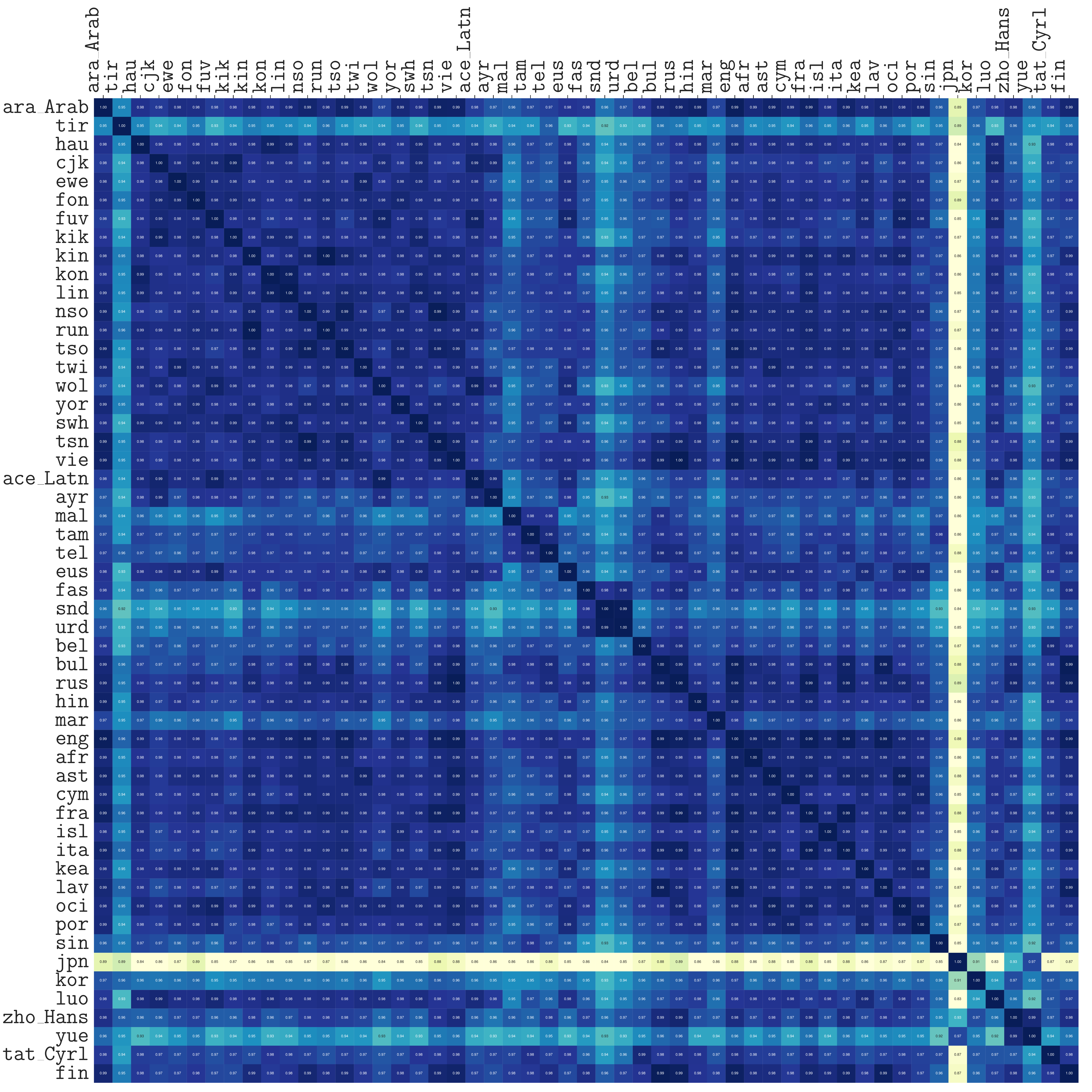}
\caption{First decoder MoE layer}\label{fig:ablation:heatmap:dec1}
\end{subfigure}
\begin{subfigure}[c]{.45\textwidth}
\includegraphics[width=\textwidth]{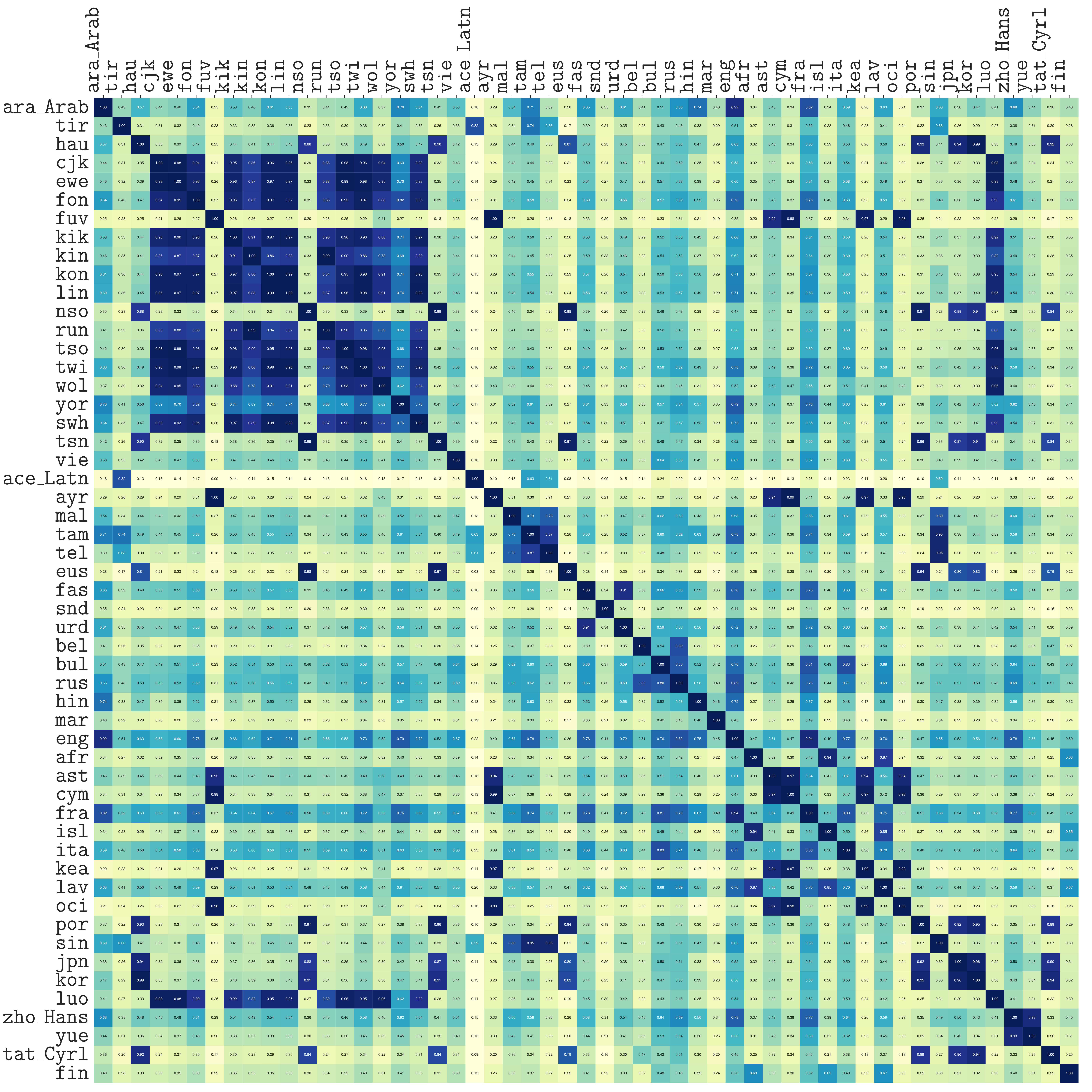}
\caption{Last decoder MoE layer}\label{fig:ablation:heatmap:dec23}
\end{subfigure}
\caption{\textbf{Cosine Similarity Scores} between languages of the \ablation~at different layers of the encoder-decoder architecture. The similarity is measured \wrt the gating decisions (expert choice) per language (source-side in the encoder and target-side in the decoder)}
\label{fig:ablation:heatmap}
\end{figure*}

We plot in \Cref{fig:ablation:heatmap} the cosine similarity scores between all 53 languages of the \ablation, in the first and last encoder MoE layer, and the first and last decoder MoE layer. 

The similarity heatmaps demonstrate that in late decoder layers (see \Cref{fig:ablation:heatmap:dec23}), the languages are being separated, i.e., dispatched to different set of experts. 
Languages within the same family are highly similar in their choice of experts, i.e.,  the late decoder MoE layers are language-specific. This is particularly the case for languages in the Atlantic-Congo family (the rows/columns from \langcode{cjk} to \langcode{yor}) and some pairs like $\{\langcode{snd\_Arab}, \langcode{urd\_Arab}\}$ in the Indo-European family or  $\{\langcode{yue\_Hant}, \langcode{zho\_Hans}\}$ in the Sino-Tibetan family.
To a lesser extent, the early encoder MoE layers (see \Cref{fig:ablation:heatmap:enc1}), also show some language-expert specialization.
The late encoder MoE layers and the early decoder MoE layers (see \Cref{fig:ablation:heatmap:enc23} and \Cref{fig:ablation:heatmap:dec1}) seem to be language-agnostic.

\subsection{Random routing}\label{app:random-routing}
\begin{figure}[!t]
\captionsetup{skip=-10pt}
\centering
\begin{tikzpicture}[scale=1]
\def\lrpair{eng-kon}
\def\hrpair{eng-fra}
\def\yminlr{10}
\def\ymaxlr{60}
\def\yminhr{1.9}
\def\ymaxhr{2.19}

\begin{groupplot}[
        group style={group size=2 by 1, horizontal sep=20pt, vertical sep=20pt},
        grid=both,
        minor tick num=1,
        label style={font=\scriptsize, yshift=-5pt},
        tick label style={font=\tiny},
        y tick label style={font=\tiny, xshift=2pt},
        x tick label style={font=\tiny, yshift=1pt, rotate=90},
        cycle list name=CustomListWithMarkers,
        height=4.5cm,
        width=\columnwidth/1.9,
        symbolic x coords={drop0.1,drop0.3,drop0.3-EOM0.1,drop0.3-FOM0.1,drop0.1-EOM0.3,drop0.1-FOM=0.3},
        xtick={drop0.1,drop0.3,drop0.3-EOM0.1,drop0.3-FOM0.1,drop0.1-EOM0.3,drop0.1-FOM=0.3},
        xticklabels={$\drop{=}0.1$,%
        $\drop{=}0.3$,$\drop{=}0.3$/$\moetokdrop{=}0.1$,$\drop{=}0.3$/$\fomdrop{=}0.1$,$\drop{=}0.1$/$\moetokdrop{=}0.3$,$\drop{=}0.1$/$\fomdrop{=}0.3$},
       ]
\nextgroupplot[
    legend style={at={(0.01,0.99)},anchor=north west, font=\tiny, nodes={scale=0.7, transform shape}},
    ylabel={\chrf{}},
   ]
    \foreach \task in {%
        en-xx-high,%
        en-xx-low,%
        xx-en-high,%
        xx-en-low,%
        non-eng-high,%
        non-eng-low%
        }{
            \addplot+[line width=1pt]
        table [y=\task,x=model,col sep=comma]{results/analysis/top2.csv};
        \addlegendentryexpanded{\task}
    }
    \legend{}

\nextgroupplot[
    legend style={at={(1.01,1.0)},anchor=north west, font=\tiny, nodes={scale=0.7, transform shape}},
    ylabel={Random-routing drop (\%)},
   ]
    \foreach \task in {%
        en-xx-high,%
        en-xx-low,%
        xx-en-high,%
        xx-en-low,%
        non-eng-high,%
        non-eng-low%
        }{
        \addplot+[line width=1pt]
        table [y=\task,x=model,col sep=comma]{results/analysis/random_drop.csv};
        \addlegendentryexpanded{\task}
    }

\end{groupplot}
\end{tikzpicture}
\caption{Performance (left) and relative drop in performance with random routing (right) of six of our MoE models.}
\label{fig:random-drop}
\end{figure}
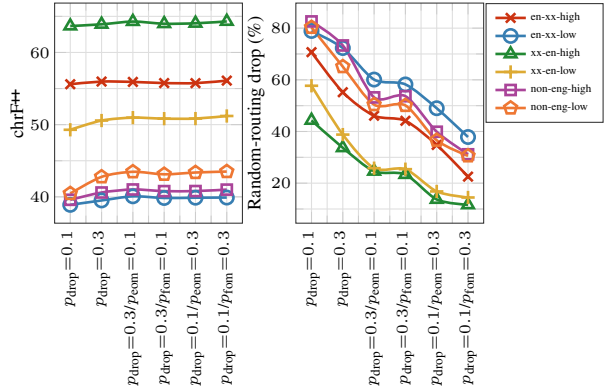
To determine how specialized are MoE experts in any given task (translation direction), we will evaluate models trained with top-2 gating
with random routing, i.e, we will choose 2 random experts instead of the top-2 experts and use their weights to compute the final output. We evaluate each model twice with different random seeds and consider the average of the two evaluation runs. The drop in translation accuracy with random routing will be a proxy of how specialized are MoE layers in any given task; if any two random experts can perform as well as the top-2, then the experts are not specialized. In \Cref{fig:random-drop} We show this drop as a percentage of the top-2 performance in chrF++ (relative chrF++ drop).

A clear and unsurprising trend we see when comparing models is that heavy regularization leads to less specialization (as shown by the diminishing relative drop in the right plot of \cref{fig:random-drop}). Non-English centric directions suffer the most from random routing followed by directions translating from English.  Translating into English suffers the least, meaning that a considerable number of decoder experts are dedicated to outputting English. Additionally, within the same subset of tasks, the low-resource directions globally suffer more from random routing.

\end{document}